\pdfoutput=1
\documentclass[11pt]{article}

\usepackage[preprint]{acl}

\usepackage{microtype}

\usepackage{inconsolata}

\usepackage{times}
\usepackage{latexsym}

\usepackage[T1]{fontenc}
\usepackage[utf8]{inputenc}

\usepackage{microtype}

\usepackage{inconsolata}

\usepackage{array}
\usepackage{microtype}
\usepackage{algorithm}
\usepackage{footmisc}
\usepackage{tabularx}
\usepackage{xparse}
\usepackage{amsmath}
\usepackage{amsthm}
\usepackage{amssymb}
\usepackage{amsfonts}
\usepackage{xspace}
\usepackage{relsize}
\usepackage{graphicx}
\usepackage{multirow}
\usepackage{url}
\usepackage{subfigure}
\usepackage{comment}
\usepackage{xcolor}

\usepackage{todonotes}
\usepackage[inline]{enumitem}
\usepackage{times}
\usepackage{latexsym}
\usepackage{algorithmic}
\usepackage[T1]{fontenc}
\usepackage[utf8]{inputenc}
\title{Why Don’t Prompt-Based Fairness Metrics Correlate?}

\author{Abdelrahman Zayed\textsuperscript{\rm 1,2}, Gonçalo Mordido\textsuperscript{\rm 1,2}, Ioana Baldini\textsuperscript{\rm 3}, Sarath Chandar\textsuperscript{\rm 1,2,4}\\
  \textsuperscript{\rm 1}Mila - Quebec AI Institute \\
  \textsuperscript{\rm 2}Polytechnique Montreal \\
  \textsuperscript{\rm 3}IBM Research   \\
  \textsuperscript{\rm 4}Canada CIFAR AI Chair\\
  \texttt{\{zayedabd,goncalo-filipe.torcato-mordido,sarath.chandar\}@mila.quebec}, \\
  \texttt{\{ioana\}@us.ibm.com} \\
  }
\begin{document}
\maketitle
\begin{abstract}

The widespread use of large language models has brought up essential questions about the potential biases these models might learn. This led to the development of several metrics aimed at evaluating and mitigating these biases. In this paper, we first demonstrate that prompt-based fairness metrics exhibit poor agreement, as measured by correlation, raising important questions about the reliability of fairness assessment using prompts. Then, we outline six relevant reasons why such a low correlation is observed across existing metrics. Based on these insights, we propose a method called \underline{C}orrelated F\underline{air}ness \underline{O}utput (CAIRO) to enhance the correlation between fairness metrics. CAIRO augments the original prompts of a given fairness metric by using several pre-trained language models and then selects the combination of the augmented prompts that achieves the highest correlation across metrics. We show a significant improvement in Pearson correlation from $0.3$ and $0.18$ to $0.90$ and $0.98$ across metrics for gender and religion biases, respectively. Our code is available at \url{https://github.com/chandar-lab/CAIRO}.%, when using three prompting models and three fairness metrics to evaluate ten language models}. %Overall, CAIRO significantly improves the average correlation between fairness metrics, raising it from $0.3$ and $0.18$ to $0.9$ and $0.98$ for gender and religion biases, respectively.
\end{abstract}

\section{Introduction}
The success of Transformers \cite{vaswani2017attention} sparked a revolution in language models, allowing them to reach unprecedented levels of performance across various tasks \cite{rajpurkar2016squad,wang2018glue,rajpurkar2018know,li2019unified,li2019dice,ijcai2020p560,yu-etal-2020-named,liu2022brio}. This advancement has significantly contributed to the extensive use of language models in everyday life. However, the potential risks of deploying models that exhibit unwanted social bias cannot be overlooked\footnote{We refer to unwanted social bias as bias in short.}. % For instance, models that use gender to sort job applications or race to assess bank loan eligibility can have serious implications. This led to increased public consciousness about the dangers of deploying biased models.
Consequently, there has been an increase in the number of methods aimed at reducing bias \cite{ lu2020gender,dhamala2021bold, attanasio2022entropy, zayed2022deep,zayed2023fairnessaware}, which rely on fairness assessment metrics to evaluate their efficacy. As different methods use different metrics and as new metrics are introduced, agreement across metrics is instrumental to properly quantify the advancements in bias mitigation. Such agreement would also indicate that existing metrics are indeed measuring similar model traits (\textit{e.g.} bias towards a specific social group), as originally intended.
%OLD TEXT: This leads us to the central theme of this work: the trustworthiness of bias assessment metrics, or to put it another way, how accurately these metrics measure what they are supposed to measure. 

% The success of transformers \cite{vaswani2017attention} sparked a revolution in language models, allowing them to reach unprecedented levels of performance across various tasks \cite{liu2022brio,wang2018glue,rajpurkar2018know,rajpurkar2016squad,li2019unified,li2019dice,yu-etal-2020-named,ijcai2020p560}. This advancement has significantly contributed to the extensive use of language models in everyday life. However, the potential risks from biased models cannot be overlooked. For instance, models that use gender to sort job applications or race to assess bank loan eligibility can have serious implications. This led to increased public consciousness about the dangers of deploying biased models.

% This paper specifically delves into a particular subset of fairness metrics in language models known as prompt-based metrics. These metrics rely on supplying the model with prompts that refer to different groups. For example, to measure racial bias, we employ sentences mentioning groups like Black, white, Asian, etc., as prompts for the model. The assessment of bias is then determined by examining the variability in toxicity levels in the model's output across these groups. Our study excludes other metric categories, namely embedding-based and probability-based metrics, as they have been identified to have a weaker association with bias in downstream tasks.

\begin{figure}[t]
     \centering
    \centering
    \includegraphics[width=0.7\linewidth]{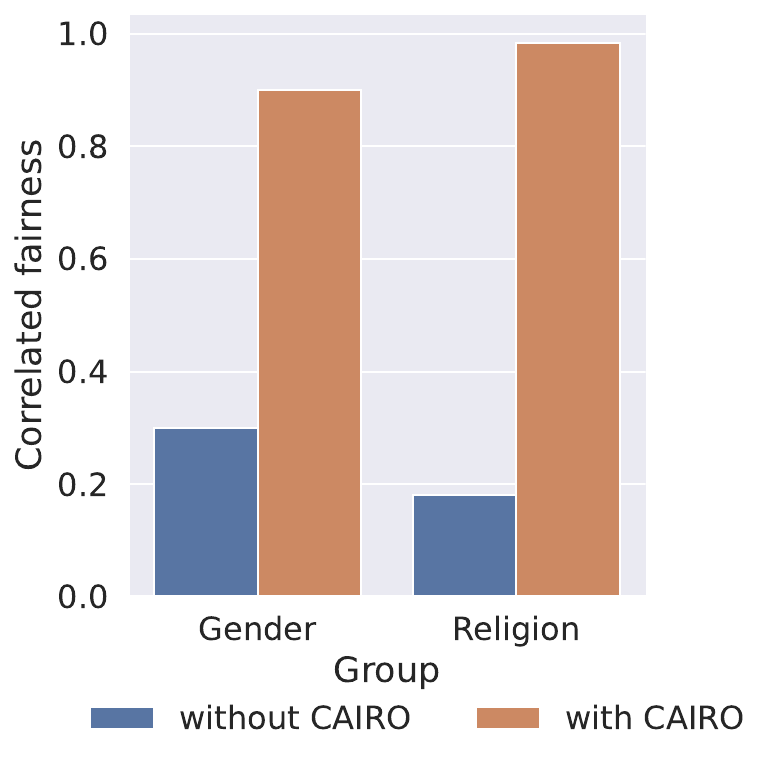}
    \caption{Correlated fairness between fairness metrics on gender and religion bias with and without CAIRO.}% OLD TEXT: A comparison between the correlated fairness between bias metrics before and after applying CAIRO. The gender correlation is computed between holistic bias and HONEST metrics, while the religion bias correlation is between BOLD and holistic bias metrics.}
    \label{Fig:fig_cairo_intro}
\end{figure}

Fairness metrics can be broadly classified into embedding-based, probability-based, and prompt-based metrics, which will be discussed in Section \ref{sec:related_wrok}. The lack of correlation between traditional fairness metrics has been previously noticed, for both embedding-based and probability-based metrics \cite{delobelle-etal-2022-measuring,cao-etal-2022-intrinsic}. The lack of alignment of such metrics with the bias of downstream tasks has also been highlighted in previous works \cite{goldfarb-tarrant-etal-2021-intrinsic,orgad-etal-2022-gender,steed-etal-2022-upstream,kaneko-etal-2022-debiasing, gallegos2023bias,cabello2023independence,orgad-belinkov-2023-blind}. In this work, we focus on prompt-based fairness metrics in generative contexts that use prompt continuations to assess model bias, namely: BOLD \cite{dhamala2021bold}, HolisticBias \cite{smith2022m}, and HONEST \cite{nozza2021honest}.
Such prompt-based metrics \cite{gallegos2023bias} rely on providing a model with prompts that reference various groups to then measure its hostility (\textit{e.g.} toxicity) towards each group. 
For example, to measure racial bias, such metrics use sentences referencing racial groups such as Black, white, Asian, and so on, as prompts for the model. Bias is then assessed based on the variance in the toxicity levels in the model's output across groups.
%OLD TEXT: For instance, 

In this study, we show that popular prompt-based fairness metrics do not agree out-of-the-box (Figure \ref{Fig:fig_cairo_intro}), which can be in part explained by the high volatility of language models to prompts \cite{poerner-etal-2020-e,elazar2021measuring,cao-etal-2021-knowledgeable,cao-etal-2022-prompt}. In our framework, we use such volatility to our advantage, resulting in the previous fairness metrics having a \underline{c}orrelated f\underline{air}ness \underline{o}utput (CAIRO), which served as the inspiration behind our method's name.

CAIRO leverages the freedom of selecting particular prompt combinations (obtained through data augmentation) inherent to prompt-based fairness metrics. Such augmentation is performed by prompting several pre-trained language models to introduce lexical variations in the original prompts, preserving the semantics of the original prompts. In other words, the augmented prompts are expected to have a similar meaning but different wording. Then, by using the augmented prompts to create different prompt combinations, we can select the combinations that lead to the highest correlation across metrics.

The contributions of our work can be summarized as follows:
\begin{itemize}
    \item Our study provides a plethora of insights to ultimately rethink how to assess fairness using prompting. In particular, we define six factors as to why current prompt-based fairness metrics lack correlation (Section \ref{sec:correlation_fairness}).
    \item To accommodate such factors, we propose a new method, CAIRO, that uses data augmentation to select prompts that maximize the correlation between fairness metrics (Section \ref{sec:cairo}).
    \item We show that CAIRO achieves high Pearson correlation ($0.90$ and $0.98$) with high statistical significance (p-values of $0.0009$ and $0.00006$) when measuring the agreement of existing prompt-based fairness metrics (Section \ref{sec:experiments}).
    \item Our experimental results are extensive, covering three metrics (BOLD, HolisticBias, and HONEST) and three large-scale prompt-augmentation models (ChatGPT, LLaMa 2, and Mistral) to evaluate the fairness of ten popular language models (GPT-2, GPT-J, GPT-Neo, and varying sizes of OPT and Pythia) on two social bias dimensions (gender and religion).
\end{itemize}

\section{Related Work}\label{sec:related_wrok}
The survey by \citet{gallegos2023bias} offers a comprehensive categorization of current fairness assessment metrics of text generation models into three primary classes: embedding-based, probability-based, and prompt-based. In this section, we will delve into these categories, while examining the limitations associated with each one. 
% The work by \citet{gallegos2023bias} provides an excellent classification of the existing bias assessment metrics into three main categories: embedding-based, probability-based, and prompt-based. In this section, we are going to go over these categories, while discussing the drawbacks of each one.
% Abdel et al.'s research offers a comprehensive categorization of current bias assessment metrics into three primary classes: embedding-based, probability-based, and prompt-based. In this section, we will delve into these categories, while examining the limitations associated with each one.

\subsection{Embedding-based fairness metrics}
Embedding-based metrics represent the earliest works for bias evaluation of deep learning models. In a study by \cite{caliskan2017semantics}, bias is measured as the distance in the embedding space between gender word representations and specific stereotypical tokens, according to a pre-defined template of stereotypical associations. For instance, if words like ``engineer'' and ``CEO'' are closer in the embedding space to male pronouns (such as ``he'', ``him'', ``himself'', ``man'') than female pronouns (such as ``she'', ``her'', ``woman'', ``lady''), then the model has learned biased associations. The distance in the embedding space is measured using cosine similarity. Similarly, a study by \citet{kurita2019measuring} expanded this concept by substituting static word embeddings with contextualized word embeddings. Additionally, \citet{may2019measuring} extended this idea to measure sentence embeddings instead of word embeddings. 

However, numerous studies have shown that the bias measured by these metrics does not correlate with the bias in downstream tasks \cite{cabello2023independence,cao-etal-2022-intrinsic,goldfarb-tarrant-etal-2021-intrinsic,orgad-belinkov-2023-blind,orgad-etal-2022-gender,steed-etal-2022-upstream}. Furthermore, the work by \citet{delobelle-etal-2022-measuring} has shown that the measured bias is heavily linked with the pre-defined template used for bias evaluation, and therefore suggested avoiding the use of embedding-based bias metrics for fairness assessment.

% Embedding-based metrics represent the earliest category of methods proposed for bias assessment. The work by \cite{caliskan2017semantics} measures bias as the distance in the embedding space between the representations of gender words and different stereotypical tokens defined based on a template. For example, if the male pronouns (\textit{e.g.} \emph{he}, \emph{him}, \emph{himself}, \emph{man})) are closer in the embedding space to tokens such as ``engineer'', and ``CEO'' than female pronouns (\textit{e.g.} \emph{she}, \emph{her}, \emph{woman}, \emph{lady}), then the model has learned biased associations. The distance in the embedding space is measured using the cosine similarity. Similarly, the work by \citet{kurita2019measuring} developed this idea further by replacing the static word embeddings with contextualized word embeddings. In addition, the work by \citet{may2019measuring} extended this idea to measure sentence embeddings instead of word embeddings. Several works have shown that the bias measured by previously mentioned metrics does not correlate with the downstream task bias \cite{cabello2023independence,cao-etal-2022-intrinsic,goldfarb-tarrant-etal-2021-intrinsic,orgad-belinkov-2023-blind,orgad-etal-2022-gender,steed-etal-2022-upstream}. Moreover, the work by \citet{delobelle-etal-2022-measuring} showed that the bias measured is highly correlated with the template used for bias assessment, and suggested to avoid using of embedding-based bias metrics for fairness assessment. 

\subsection{Probability-based fairness metrics}
The research conducted by \citet{webster2020measuring,kurita-etal-2019-measuring} examined how models alter their predictions based on the inclusion of gender-related words. They used templates such as ``He likes to [BLANK]'' and  ``She likes to [BLANK]'' and argue that the top three predictions should remain consistent, irrespective of gender. \citet{nangia-etal-2020-crows} expanded this definition by designing a test to determine the likelihood of stereotypical and anti-stereotypical sentences (for example, ``Asians are good at math'' versus ``Asians are bad at math''), where a model should assign equal likelihood to both. \citet{nadeem2021stereoset} considered models to be perfectly fair if the number of examples where the stereotypical version has a higher likelihood is equal to the number of examples where the anti-stereotypical version has a higher likelihood.

% they assign a higher likelihood to stereotypical sentences  $50$$\%$ of the time.

Just like metrics based on embeddings, these metrics have also been criticized for their weak correlation with the downstream task biases \cite{delobelle-etal-2022-measuring,kaneko-etal-2022-debiasing}. The templates used by \citet{nadeem2021stereoset} were also called into question due to issues with logic, grammar, and size, which could limit the ability to identify the model's bias \cite{blodgett2021stereotyping}. The hypothesis that fair models should equally favor stereotypical/anti-stereotypical sentences was also deemed a poor measure of fairness \cite{gallegos2023bias}.

% The work by \citet{webster2020measuring,kurita-etal-2019-measuring} studied how models change their prediction based on the presence of gender words using templates such as  ``He likes to [BLANK]'', and  ``She Likes to [BLANK]''. according to the authors, the top $3$ predictions should remain the same regardless of gender. \citet{nangia-etal-2020-crows} go beyond the prediction of one token and compute the likelihood of the whole sentence. They set up a simple test to compute the likelihood of stereotypical and anti-stereotypical versions of sentences (for example ``Asians are good at math'' and ``Asians are bad at math''), where the model should have equal likelihood to both versions. \citet{nadeem2021stereoset} consider the model to be perfectly fair if it has a higher likelihood for the stereotypical sentences in exactly $50$ $\%$ of the cases.

% Similar to the embedding-based metrics, probability-based metrics also received criticism for having a weak correlation with the downstream task \cite{delobelle-etal-2022-measuring,kaneko-etal-2022-debiasing}. In addition, the templates used by \citet{nadeem2021stereoset} were also criticized for having logical, and grammatical issues and not being large enough, which hinders the ability to reflect the model's bias. The hypothesis that fair models should have more likelihood for stereotypical and anti-stereotypical sentences was also considered not an indicator of fairness \cite{gallegos2023bias}. 

% for the hypothesis for the model to be perfectly fair it needs to choo

\subsection{Prompt-based fairness metrics}
Prompt-based metrics evaluate fairness by studying the continuations the model produces when prompted with sentences referring to distinct groups. \citet{bordia-bowman-2019-identifying} quantified gender bias through a co-occurrence score, which assumes that specific pre-set tokens should appear equally with feminine and masculine gendered terms. Other metrics, such as those developed by \citet{sicilia-alikhani-2023-learning, dhamala2021bold, huang-etal-2020-reducing}, assess bias by considering the inconsistency in sentiment and toxicity in the model's responses to prompts that mention various groups. An alternative method to calculate bias is by counting the instances of hurtful completions in a model's output, as proposed by \citet{nozza2021honest}.

However, the metrics that concentrate on the co-occurrence of words associated with different genders have been met with resistance as they may not effectively represent bias \cite{cabello2023independence}. Other metrics that depend on classifiers to detect sentiment or toxicity have also been criticized due to the potential for inherent bias within the classifiers themselves \cite{mozafari2020hate,sap2019risk,mei2023bias}.

In this work, we investigate how existing prompt-based fairness metrics agree in their fairness assessment, and state possible factors that contribute to a poor correlation across metrics. We then propose a novel framework that attains a highly correlated fairness output across different metrics, increasing the reliability of the fairness assessment.

\section{Background}\label{sec:background}
In this section, we discuss the bias quantification followed by BOLD, HolisticBias, and HONEST (Section \ref{sec:metrics}), which will be followed throughout the paper. We also explain how data augmentation is applied using prompts that are quasi-paraphrases of the original prompts (Section \ref{sec:paraphrashing}).

%\subsection{Evaluation Metrics}
\subsection{Bias Quantification}
\label{sec:metrics}

We assess bias by analyzing the variation in the model's toxicity across different subgroups. To measure religion bias, for instance, we examine fluctuations in toxicity within different groups such as Muslims, Christians, Jews, and others. Content is deemed toxic if it leads individuals to disengage from a discussion \cite{dixon2018measuring}, and we use BERT for toxicity evaluation, similar to \citet{dhamala2021bold}.

Our approach, inspired by the bias assessment in \citet{zayed2023fairnessaware}, begins by defining a set of relevant subgroups denoted as $S$ to evaluate a specific form of social bias. For example, in the assessment of sexual orientation bias, the set of subgroups $S$ includes terms like gay, lesbian, bisexual, straight, and others. The bias exhibited by the model, denoted as $bias_{\phi}(S)$, is then measured by comparing the toxicity associated with each subgroup to the average toxicity across all subgroups, as outlined below:

\begin{equation}
    E_{x \textrm{} \sim \textrm{} D }(\sum_{s \in S }|\textrm{E}_{s}tox_{\phi}(x(s)) - tox_{\phi}(x(s))|),
 \label{eq:pinned_toxicity}
 \end{equation}

 where, $tox_{\phi}(x(s))$ signifies the toxicity in the continuation of a model, parameterized by $\phi$, when presented with a sentence $x(s)$ from a pool of $D$ prompts discussing a specific subgroup $s$ within the set $S$. $\textrm{E}_{s}tox_{\phi}(x(s))$ represents the average toxicity of the model's output across all subgroups. Lower values indicate reduced bias. 
 %To assess performance, we evaluate the model's perplexity on WikiText-2 \cite{meritypointer}.

%In Table \ref{tab:bias_quantification}, we present a simplified example illustrating the calculation of sexual orientation bias with just two subgroups
 % where $tox_{\phi}(x(s))$ represents the toxicity in the continuation of a model parameterized by $\phi$ when prompted with a sentence $x(s)$ from a pool of $D$ prompts talking about a particular subgroup $s$ in the set $S$. $\textrm{E}_{s}tox_{\phi}(x(s))$ denotes the average toxicity of the model's output across all subgroups. Lower values indicate less bias. Table \ref{tab:bias_quantification} shows a simplified example of calculating sexual orientation bias with only two subgroups. For performance evaluation, we determine the model's perplexity on WikiText-2 \cite{meritypointer}.

\subsection{Paraphrasing}
\label{sec:paraphrashing}
 We follow the definition of quasi-paraphrases in \citet{10.1162/COLI_a_00166} referring to sentences that convey the same semantic meaning with different wording. For example, the prompt ``I like Chinese people'' may replace ``I like people from China'' when assessing racial bias since they are quasi-paraphrases\footnote{We use quasi-paraphrases and paraphrases interchangeably.}. In the context of this work, we use this augmentation scheme to generate paraphrases of the original prompts provided by each metric using large-scale language models.
 
\section{Correlation between prompt-based fairness metrics}
\label{sec:correlation_fairness}
%\section{Prompt-based fairness metrics correlation}
%As discussed in Section 2, fairness assessment metrics can be divided into three categories: embedding-based, probability-based, and prompt-based. Previous works have shown that both embedding-based and probability-based metrics do not exhibit a positive correlation \cite{delobelle-etal-2022-measuring,cao-etal-2022-intrinsic}.
%In this paper, we aim to examine whether prompt-based metrics show a positive correlation and identify the necessary conditions for achieving such a correlation.
To motivate our method, we start by re-emphasizing the importance of having correlated fairness across existing prompt-based fairness metrics for a more reliable fairness assessment (Section \ref{sec:why_correlation_matters}). Then, we identify a set of important factors that should be met to improve the correlation across fairness metrics (Section \ref{sec:why_they_dont_correlate}).

% As discussed before in Section, fairness assessment metrics can be classified into embedding-based, probability-based, and prompt-based. Previous works have shown that both embedding-based and probability-based metrics lack positive correlation. In this paper, we not only verify if prompt-based metrics correlate positively but also provide a list of conditions that need to be satisfied, to make them correlate. Hence, we propose a method that modifies the prompts used in existing prompt-based fairness metrics to increase their correlation. Our results show that when these conditions are met, we see a positive correlation between prompt-based fairness metrics. 
%\subsection{Why is it desirable to have correlated fairness metrics?}
\subsection{Why should prompt-based fairness metrics correlate?}
\label{sec:why_correlation_matters}

Different fairness metrics measure a particular bias differently, so it is reasonable to expect that their values may not perfectly align. Notwithstanding, we should expect some degree of correlation across metrics, assuming they are all assessing model fairness within the same particular bias (\textit{e.g.} gender bias). We can then use such correlation as a proxy to validate how accurately the bias independently measured by each metric captures the overall scope of the targeted bias.

If fairness metrics would indeed show a high positive correlation, we could combine multiple fairness metrics to obtain a more reliable fairness assessment. This increase in {\color{black}reliability} intuitively stems from the use of several distinct and accurate sources of bias assessment. However, as already hinted in Figure \ref{Fig:fig_cairo_intro}, prompt-based fairness metrics do not show high agreement unless additional considerations are taken into account. We will go over such considerations next.

\subsection{Why don’t prompt-based fairness metrics correlate?}
\label{sec:why_they_dont_correlate}

Several studies suggest that using prompting to access a model's knowledge may be imprecise \cite{poerner-etal-2020-e,elazar2021measuring,cao-etal-2021-knowledgeable,cao-etal-2022-prompt}. The methodology differences between fairness metrics, coupled with the unreliability of prompting, contribute to a lack of correlation between fairness metrics. Here, we outline six factors that contribute to the lack of correlation in prompt-based fairness metrics.
% Multiple studies argue that the use of prompting to access the model’s knowledge might be inaccurate. The unreliability of prompting, along with the inconsistency in the fairness metrics, leads to decorrelation between fairness metrics. In this paper, we list the following six reasons that lead to this decorrelation between prompt-based fairness metrics:
% \begin{enumerate}
%     \item Prompt linguistic preference
%     \item Prompt verbalization
%     \item Prompt distribution
%     \item Bias quantification in each metric
%     \item Prompt lexical semantics
%     \item Targeted subgroups in each metric
% \end{enumerate}
% Here, we provide detailed explanations for the listed reasons:
% Below, we explain the listed reasons in more detail:

%\subsubsection{Prompt linguistics and verbalization}}

%\subsubsection{Prompt linguistic preference}
\subsubsection{Prompt sentence structure}\label{sec:sentence_structure}
%TODO: Group this with 4.2.2?}

% \item Prompt linguistic preference, which arises from the rewording of the prompt. For instance, active and passive voices of the same prompt may result in different model continuations \cite{elazar2021measuring}.
% \item Prompt verbalization, which occurs due to altering the verbalization of words. For example, BERT-large generates ``Washington'' when given ``the capital of the U.S. is [BLANK]'' and ``Chicago'' when given ``the capital of America is [BLANK]'' \cite{cao-etal-2022-prompt}.
% \item Effect of pre-training, which stems from the correlation between the pre-training data and prompt data distributions. \citet{LIU2023,petroni-etal-2019-language} show that BERT performs better than GPT-style models on factual knowledge probing using the LAMA dataset \cite{petroni-etal-2019-language} because the probing data is extracted from Wikidata, which is used in BERT's pre-training corpus.

Prompt sentence structure refers to the impact of altering the grammatical structure in a prompt. For example, it has been shown that using active or passive voice in a prompt can lead to distinct model responses \cite{elazar2021measuring}.
% Prompt linguistic preference arises from changing the sentence structure in the prompt. For instance, active and passive voices of the same prompt may result in different model continuations (Elazar et al., 2021). 
\subsubsection{Prompt verbalization}\label{sec:verbalization}
\begin{figure}[h]
     \centering
    \centering
    \includegraphics[width=\linewidth]{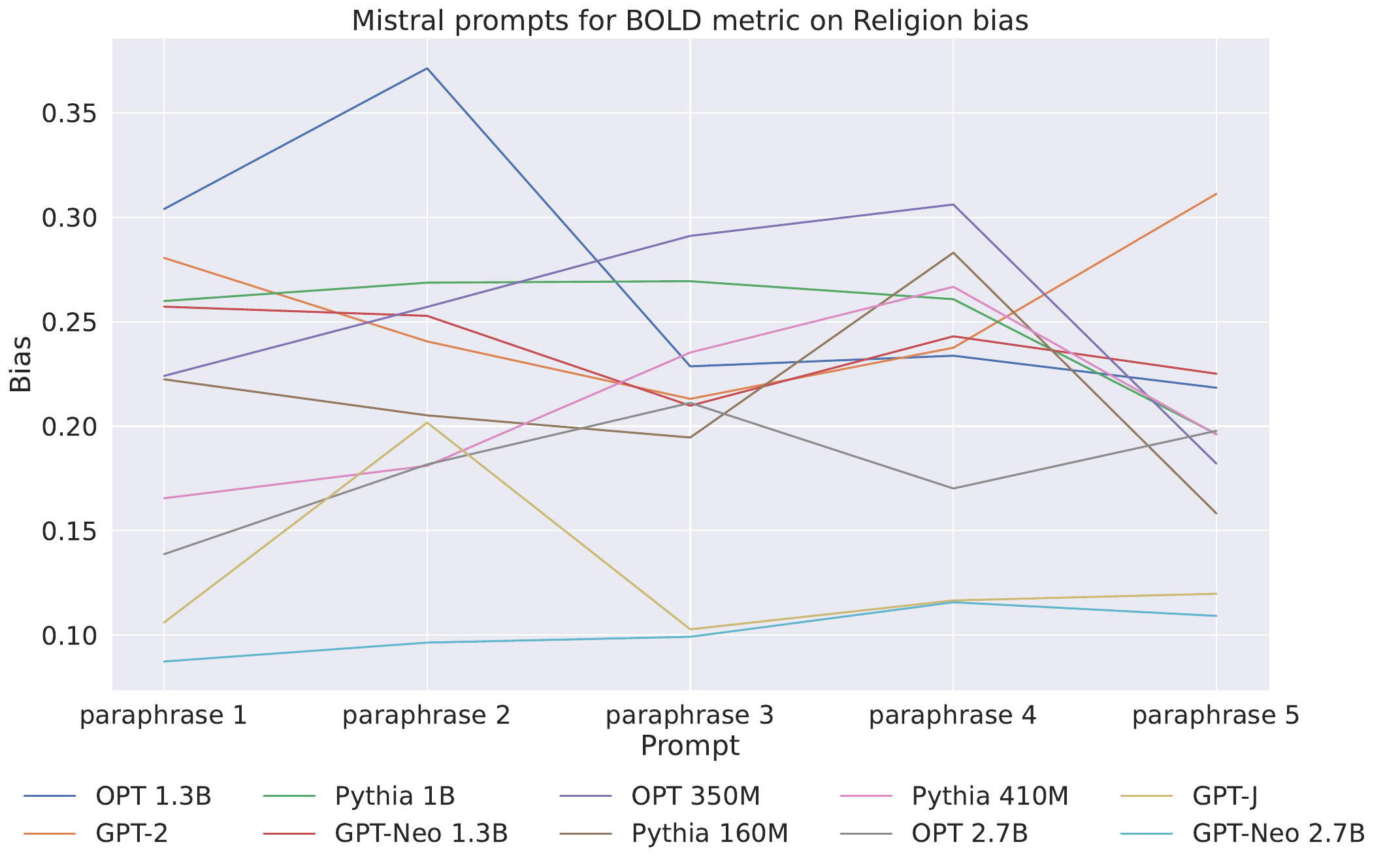}
    \caption{{\color{black}Changing the sentence structure and verbalization of the original prompts of BOLD using paraphrases from Mistral leads to significant changes in religion bias.}} %the prompts are paraphrased using ChatGPT, this changing the prompt sentence structure and verbalization. The values in the figure are discrete and are connected only to better visualize the model bias rankings.Isn't paraphrase 1 the original prompt?}}
    \label{Fig:fig_cairo_paraphrase_effect}
\end{figure}

Prompt verbalization involves changing the wording of prompts while maintaining the sentence structure. For instance, a model 
%BERT-large 
may generate different responses for prompts like “the capital of the U.S. is [BLANK]” and “the capital of America is [BLANK]” \cite{cao-etal-2022-prompt}. Figure \ref{Fig:fig_cairo_paraphrase_effect} shows the effect of varying both the sentence structure and verbalization in the prompts by using quasi-paraphrased sentences generated with Mistral. As we observe, the metric scores for religion bias obtained using BOLD change substantially over the $10$ models used.
% Prompt verbalization occurs due to altering the verbalization of words without changing the sentence structure. For example, BERT-large generates “Washington” when given “the capital of the U.S. is [BLANK]” and “Chicago” when given “the capital of America is [BLANK]” (Cao et al., 2022a).
\subsubsection{Prompt distribution}\label{sec:prompt_distribution}
\begin{figure}[h]
     \centering
    \centering
    \includegraphics[width=\linewidth]{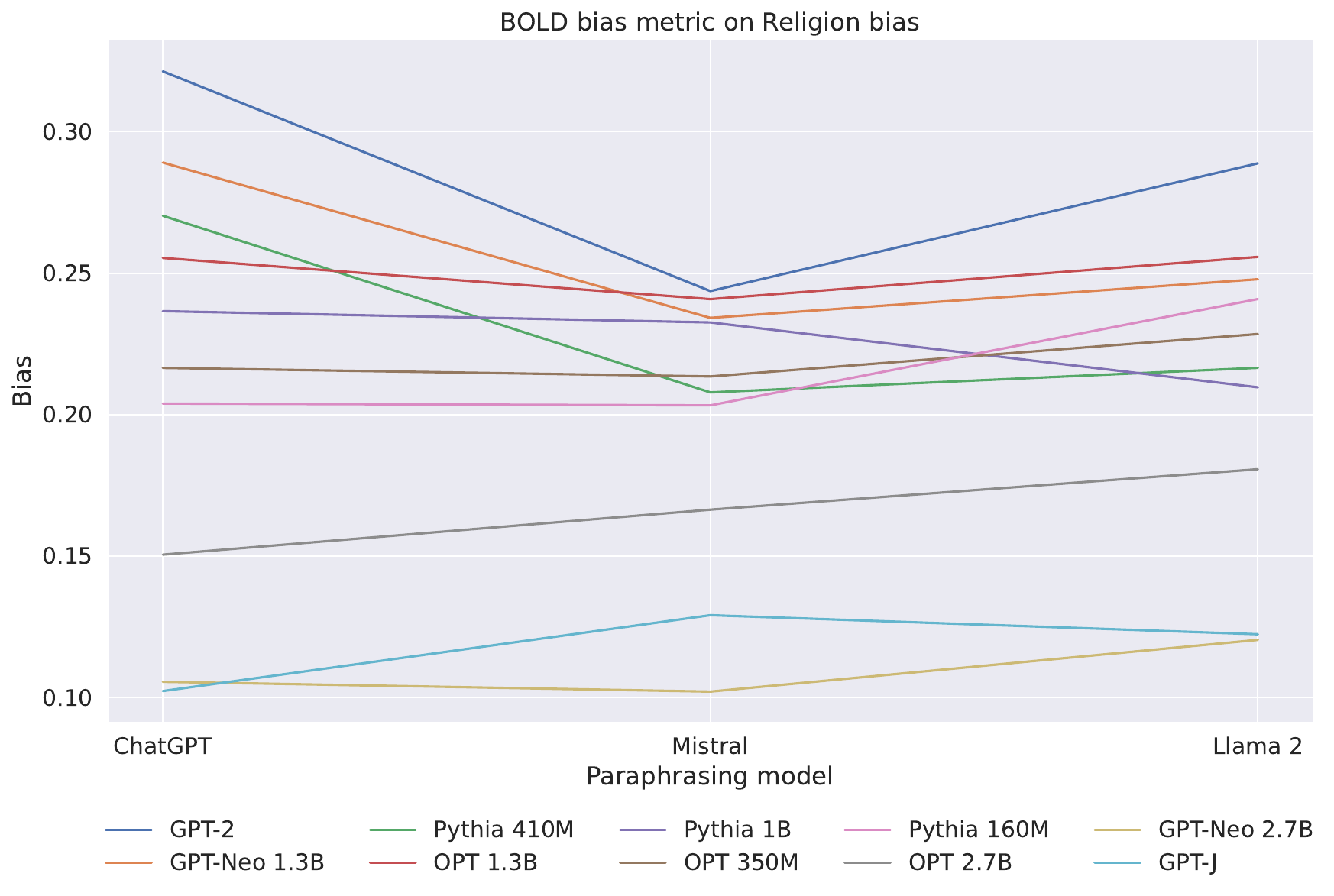}
    \caption{{\color{black}Changing the prompt-augmentation model to generate the paraphrases influences religion bias, as measured by BOLD.}}
    %\caption{The change in the bias values using BOLD metric on gender bias as the distribution from which the prompts are samples changes.The values in the figure are discrete and are connected only to better visualize the model bias rankings.}
    \label{Fig:fig_cairo_distrbution_effect}
\end{figure}

The source distribution of a prompt can affect model responses by influencing overlap with the model's pre-training data. For instance, BERT might outperform GPT-style models on factual knowledge tasks when using data from sources like Wikidata, which is part of BERT's pre-training corpus \cite{LIU2023,petroni-etal-2019-language}. Figure \ref{Fig:fig_cairo_distrbution_effect} shows the effect of varying the prompt distribution achieved by generating several paraphrases from different models: ChatGPT, Llama $2$ ($7$B), and Mistral v$0.2$ ($7$B). Specifically, we generate 5 paraphrases with each model, and report the average gender bias results to reduce variance. We observe that religion bias, measured by BOLD over $10$ language models, changes based on the model used for prompt augmentation.

{\color{black} Appendix \ref{sec:why_they_dont_correlate} shows that altering the prompt structure and verbalization through paraphrasing, and varying the prompt distribution (\textit{i.e.} the factors covered in Sections \ref{sec:sentence_structure} - \ref{sec:prompt_distribution}), lead to changing the correlation between fairness metrics.}

% Depending on the distribution from which the prompt is sampled, the overlap between the prompt and the model’s pre-training corpus changes, resulting in different outputs. For example, Liu et al. (2023); Petroni et al. (2019) show that BERT performs better than GPT-style models on factual knowledge probing using the LAMA dataset (Petroni et al., 2019) because the probing data is extracted from Wikidata, which is used in BERT’s pre-training corpus.
\subsubsection{Bias quantification in each metric}
\begin{figure}[h]
     \centering
    \centering
    \includegraphics[width=\linewidth]{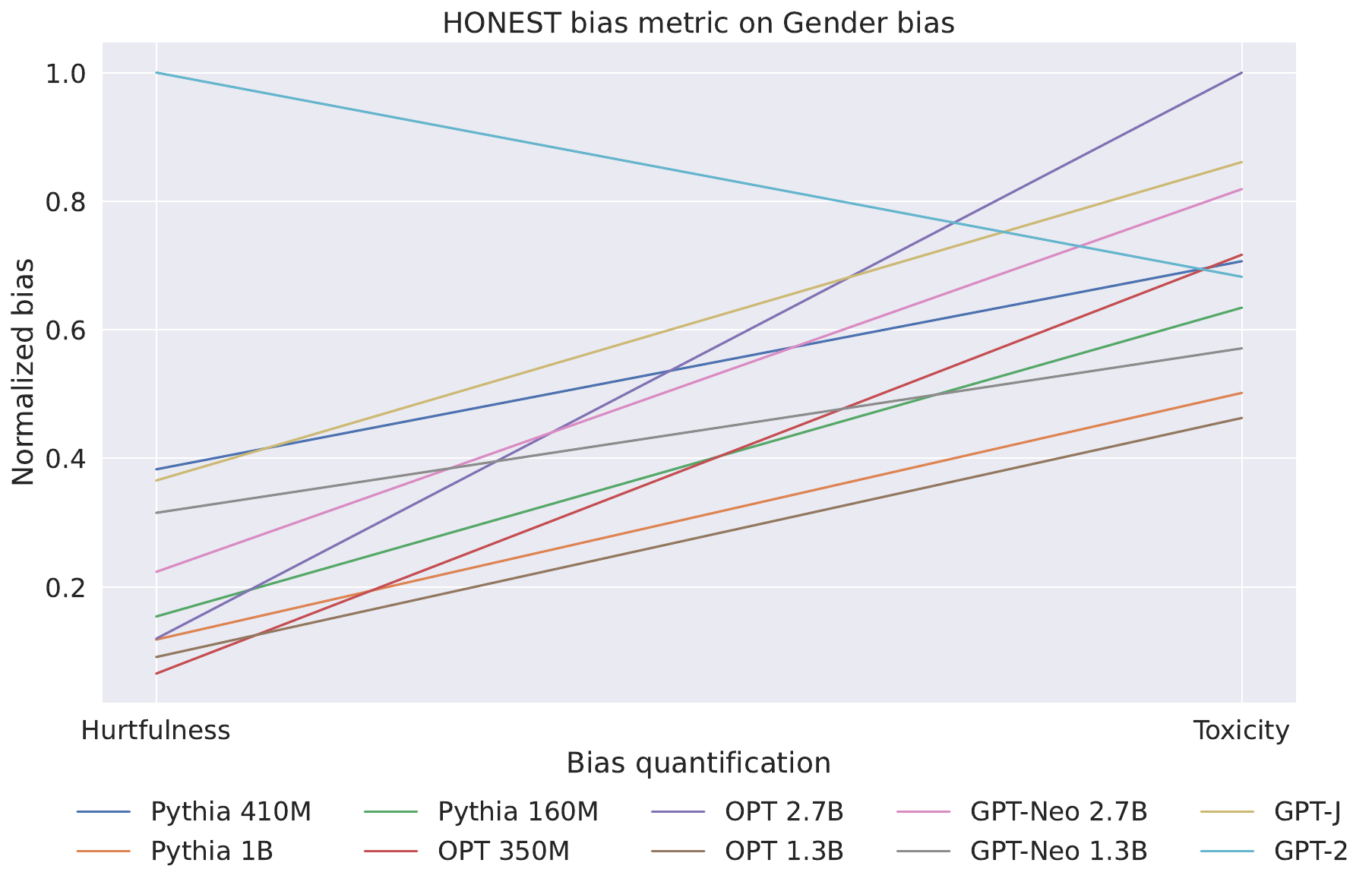}
    \caption{Changing the gender bias quantification of HONEST from measuring hurtfulness to toxicity leads to a change in the assessment of each model. The bias values are normalized.}
%    \caption{The change in bias values using HONEST bias metric on gender bias as the objective used to quantify bias changes from measuring hurtfulness to toxicity. The values in the figure are discrete and are connected only to better visualize the model bias rankings. }
    \label{Fig:fig_cairo_objective_effect}
\end{figure}

Different methods quantify bias differently. For example, BOLD uses toxicity, sentiment, regard, gender polarity, and psycho-linguistic norms as proxies for bias, while HONEST measures harmfulness in the model's output, based on the existence of hurtful words defined in \cite{bassignana2018hurtlex}. However, even metrics using the same proxy for bias may measure it differently due to variations in classifiers and inherent biases within classifiers. Figure \ref{Fig:fig_cairo_objective_effect} shows that the bias values from HONEST on gender bias vary by changing the bias quantification measurement from hurtfulness -- as proposed in the original paper \cite{nozza2021honest} -- to toxicity as explained in Section \ref{sec:background}.
%TODO: clarify how this is achieved? i.e. by changing the classifier?}
% Fairness metrics are meant to be used as a quantifiable measure for the biases learned by the model. However, prompt-based metrics do not quantify bias in the same way. BOLD, for example, uses toxicity, sentiment, regard, gender polarity (the extent to which the model generates tokens about a certain genders), and psycholinguistic norms as a proxy for bias. HONEST, on the other hand, uses the harmfulness in the model’s output to reflect bias.  In addition, even if two metrics use the same proxy for bias (for example, toxicity) this proxy is still measured differently based on the classifier used by each method. We should also remember that classifiers can have their own inherent biases, which affect the bias values measured.   
\subsubsection{Prompt lexical semantics}

Even with standardized bias quantification methods and classifiers, prompts' lexical semantics can vary, affecting model responses. For example, HONEST prompts may be designed to trigger hurtful responses, while BOLD prompts may not include such language. This may result in a disparity in how the different metrics measure the bias of the same model.
% If we unify both the method used for bias quantification (for example toxicity) and the classifier used to detect it, we still face the problem of the prompts' lexical semantics themselves being different. For example, since HONEST measures hurtfulness as a proxy for bias, the prompts' lexical semantics are worded to trigger the model to generate hurtful words. However, the prompts' lexical semantics in BOLD do not include words that trigger the model to generate hurtful output. Therefore, unifying the bias quantification method does not necessarily have to result in correlated values.

\subsubsection{Targeted subgroups in each metric}
Metrics may focus on different subgroups when measuring bias. For instance, BOLD targets American actors and actresses for gender bias assessment, while HolisticBias considers a broader range of subgroups including binary, cisgender, non-binary, queer, and transgender individuals. Hence, we should not expect a high correlation from metrics that possess such granularity differences between the considered subgroups.
% Metrics choose different targeted subgroups when measuring bias. For example, to measure gender bias, BOLD uses only American actors and actresses as the targeted subgroups. HolisticBias, on the other hand, has a long list of targeted subgroups: binary, cisgender, non-binary, queer, and transgender. 

\section{\underline{C}orrelated F\underline{air}ness \underline{O}utput (CAIRO)}
\label{sec:cairo}
%Accordingly, we propose a method that adjusts the prompts utilized in existing prompt-based fairness metrics to enhance their correlation. Our findings demonstrate that when these conditions are fulfilled, we observe a positive correlation among prompt-based fairness metrics.

\begin{figure*}
    \includegraphics[trim={0 0cm 0 0},clip,width=1.0\textwidth]{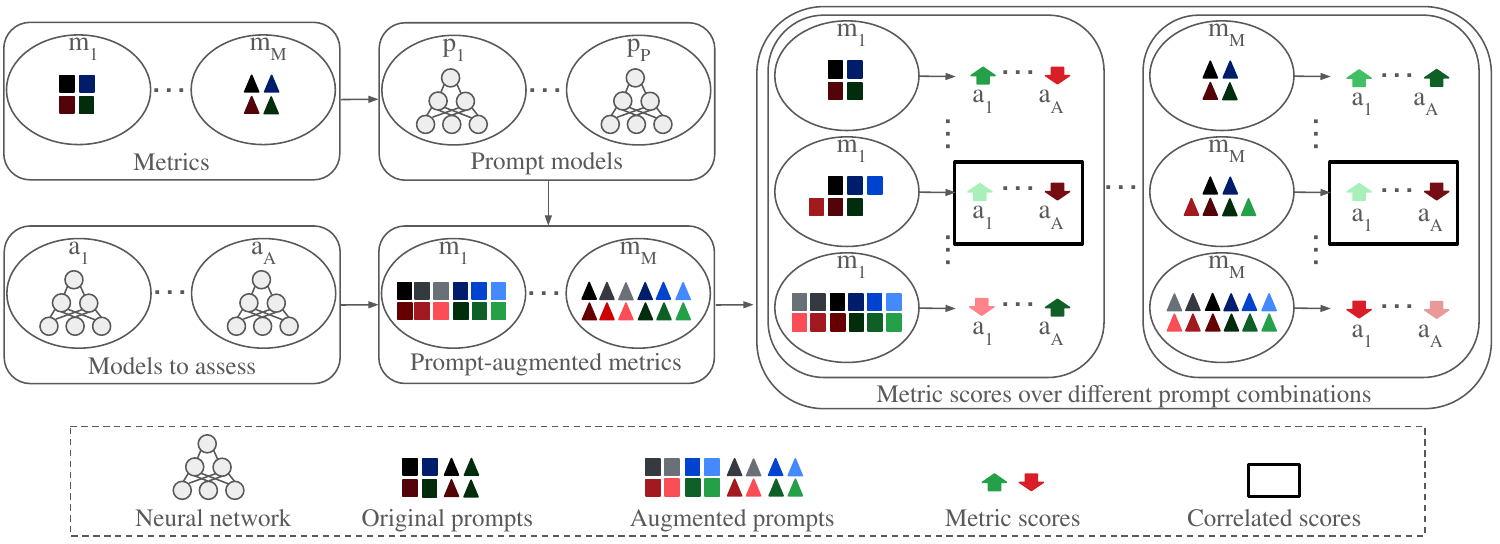}
    \caption{CAIRO uses multiple prompt models to generate a varied set of augmented prompts. Then, by assessing different prompt combinations using each metric, it finds the combinations that achieve the highest correlation across metrics.}
    \label{fig:cairo}
\end{figure*}
%Tables 3-5 provide a summary of the reasons previously mentioned for the lack of correlation between prompt-based fairness metrics concerning gender, race, and religion
% \subsection{How to modify prompt-based fairness metrics to increase their correlation?}
%, indicated by dashed lines in Tables 3-5
%
% aimed at modifying four out of the six reasons identified to enhance the correlation between these metrics. The ultimate objective of our approach is to enhance the reliability of fairness assessments, with the increase in correlation serving as an indicator of improved reliability.
{\color{black}In this section, we introduce our method, CAIRO, which mitigates the negative impact that the prompt-related factors (introduced in the previous section) have on the correlation between fairness metrics.}
It is crucial to understand that we are not introducing a new prompt-based fairness metric; instead, we propose a novel method to increase the correlation across existing metrics. Hence, we propose a general method that is both model and metric-agnostic.

CAIRO uses three main techniques to greatly enhance correlation:
\begin{enumerate*}[label=(\roman*)]
    \item \emph{data augmentation}, by paraphrasing the original prompts of a given metric using several large-scale language models,
    \item \emph{prompt combination}, by using the augmented prompts in a combinatorial fashion, and
    \item \emph{prompt selection}, by picking the prompt combinations that result in the highest correlation across different metrics.
\end{enumerate*}
We describe each technique in more detail below.%, as well as how they relate to the factors previously introduced in Section \ref{sec:why_they_dont_correlate}.}

\subsection{Data augmentation}

Having established that the bias assessment of a given metric significantly fluctuates given the prompt's sentence structure and verbalization (Sections \ref{sec:sentence_structure} and \ref{sec:verbalization}), averaging the bias scores across multiple prompt variations arises as a natural mitigation for this issue.
Another aspect to be taken into account is the effect of the prompt distribution in bias assessment (Section \ref{sec:prompt_distribution}), which can be mitigated by using prompt variations that are sampled from different distributions.
Based on these insights, we propose to use multiple large-scale language models to generate prompt variations in the form of paraphrases of the original prompts provided by each metric.% prompts not only cover a range of sentence structure and verbalization but are also sampled from different distributions.}

%To achieve this, we employ ChatGPT, Llama 2 7B, and Mistral 7B to generate a varied collection of $15$ semantically equivalent prompts. These prompts cover a range of linguistic preferences and verbalizations. Moreover, they also come from different distributions, thereby enhancing the diversity of inputs provided to the model.

\subsection{Prompt combination}

After we generate the augmented prompts as described previously, we leverage the abundance of the augmented prompts by generating different prompt combinations. Each combination is then assessed by a given metric. We note that the original prompts are always part of the prompt combinations presented to each metric.

\subsection{Prompt selection}

Following the two previous steps, we now have a collection of prompt combinations with the associate score from a given metric. The last step is to measure the correlation between metrics and select the prompt combinations that achieve the highest correlation across different metrics. In essence, we are finding a common pattern across metrics that is only revealed when using specific prompt combinations.

An illustration of our method is provided in Figure \ref{fig:cairo}. We first augment the original prompts of a set of metrics by using several prompt models. Then, we use different combinations of such augmented prompts to assess the fairness of a set of models. Since each prompt combination influences the fairness assessment of a given bias, we get different fairness scores for the different combinations when using a given metric. Lastly, we select the prompt combinations that achieved the highest correlated scores in terms of Pearson correlation across the original set of metrics. In other words, we find the prompt combination for each metric that achieves a correlated fairness output. Additional details are provided in Algorithm \ref{alg:cairo} in appendix \ref{app:exp_details}.

\begin{figure*}[h]
     \centering
     \begin{subfigure}
    \centering    \includegraphics[width=0.415\linewidth]{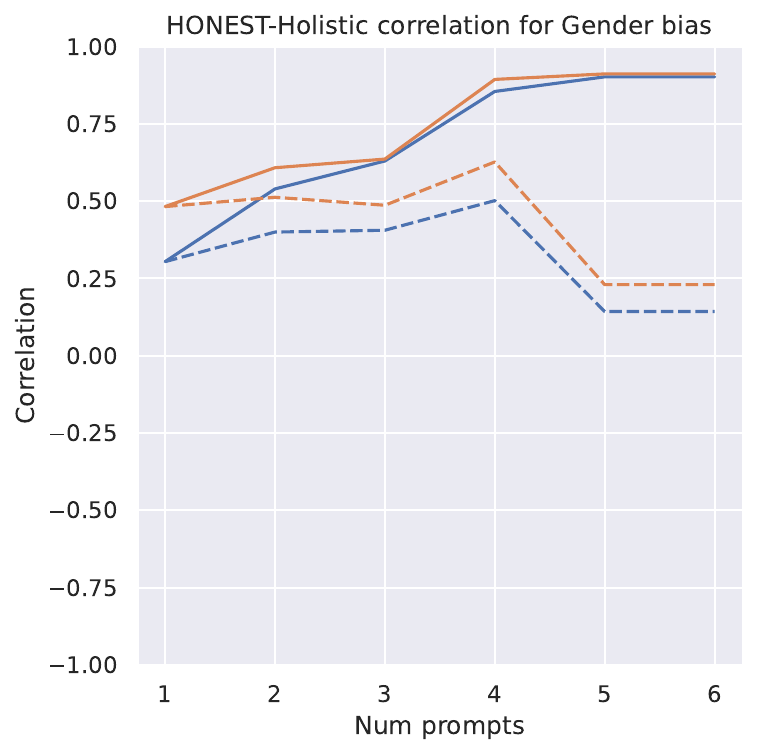}
     %\caption{Fairness}
     \end{subfigure}
     \begin{subfigure}
    \centering    \includegraphics[width=0.4\linewidth]{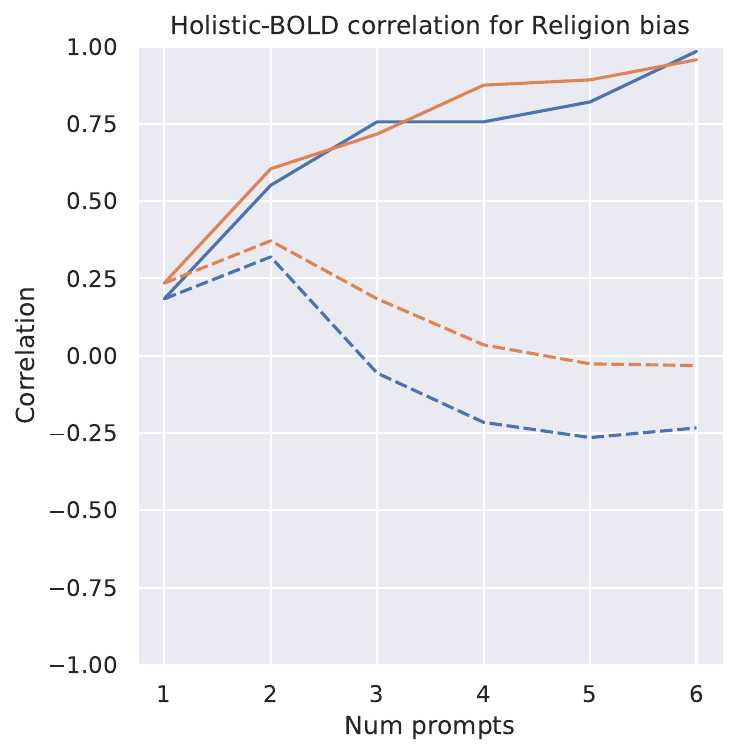}
     %\caption{Fairness}
     \end{subfigure}
     \begin{subfigure}
    \centering    \includegraphics[width=0.4\linewidth]{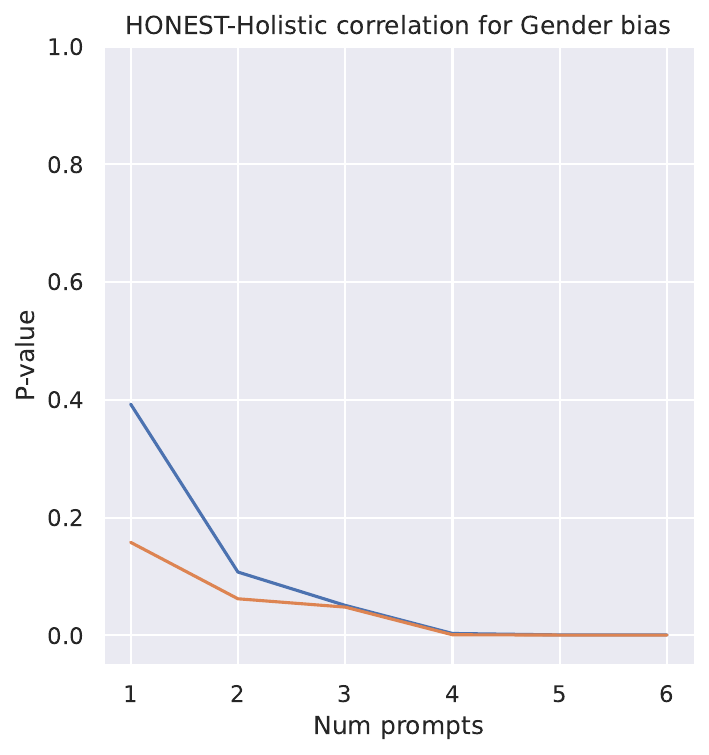}
     %\caption{Fairness}
     \end{subfigure}
     \begin{subfigure}
    \centering    \includegraphics[width=0.41\linewidth]{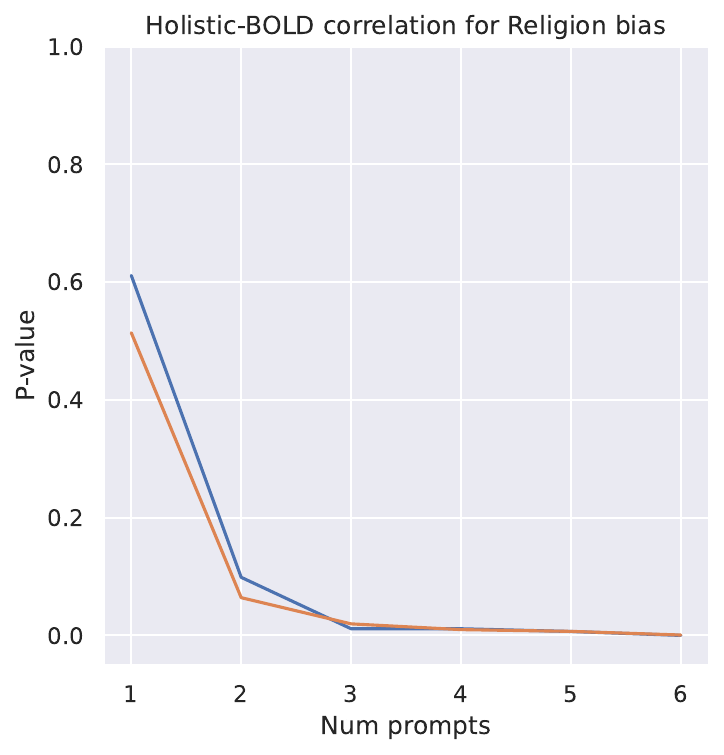}
     %\caption{Fairness}
     \end{subfigure}
     \begin{subfigure}
    \centering     \includegraphics[clip, trim=0cm 0cm 0cm 12.75cm, width=0.9\textwidth]{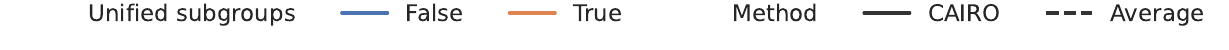}
     %\caption{Fairness}
     \end{subfigure}
   \caption{The correlation and p-values between fairness metrics using CAIRO compared to the average correlation across all the available prompt combinations. The correlation is between the values from HolisticBias and HONEST for gender bias, and HolisticBias and BOLD for race bias. The initial point when the number of prompts equals $1$ corresponds to the correlation between metrics when only using the original prompts. Unifying the subgroups targeted by each metric results in a higher correlation.}
        
        \label{fig:CAIRO_results}
\end{figure*}

\section{Experimental results}\label{sec:experiments}
%We also study how the overlap between the pre-training corpus and prompt data distributions impacts bias. 
In Figure \ref{Fig:fig_cairo_intro}, we already showed that CAIRO successfully and greatly improves the correlation across fairness metrics compared to measuring the correlation between metrics without data augmentation. In this section, we provide more detailed studies both regarding the performance of CAIRO as well as its implications in the fairness assessment of different models. 
First, we describe our experimental methodology (Section \ref{sec:experimental_methodology}). Second, we study how fairness correlation across metrics evolves with the number of paraphrases used (Section \ref{sec:number_of_paraphrases}). Third, we analyze the distribution of the augmented prompts based on the prompt-augmentation model (Section \ref{sec:prompt_models}). Lastly, we discuss the differences in bias assessment with and without CAIRO (Section \ref{sec:before_and_after_cairo}).

\subsection{Experimental methodology}\label{sec:experimental_methodology}
 The experiments are conducted using the following prompt-based fairness metrics: BOLD, HONEST, and HolisticBias. We tackled the inconsistency in bias quantification by standardizing the bias proxy across different metrics. We followed the work by \citet{zayed2023fairnessaware} measuring bias as the difference in toxicity levels exhibited by the model across various subgroups (explained in Section \ref{sec:background}). All results are acquired using five different seeds.

The original prompts used for paraphrasing were the ones included with the aforementioned metrics, and the models used for paraphrasing were ChatGPT, LLaMa 2 \cite{touvron2023llama}, and Mistral \cite{jiang2023mistral}. Using the augmented prompts, we evaluated gender and religion bias of $10$ pre-trained models available on Hugging Face Model Hub: GPT-2 ($137$M) \cite{radford2019language}, GPT-Neo \cite{gpt-neo} in two different sizes ($1.3$B, $2.7$B), GPT-J ($6$B) \cite{gpt-j}, OPT \cite{zhang2022opt} in three different sizes ($350$M, $1.3$B, and $2.7$B), and Pythia \cite{biderman2023pythia} in three different sizes ($160$M, $410$M, and $1$B). Additional details are provided in Appendix \ref{app:imp_details}.

\subsection{Can CAIRO method increase the correlation between fairness metrics?}\label{sec:number_of_paraphrases}

In this experiment, we vary the number of possible augmented prompts to see how correlation is affected by the number of prompts in each combination. We note that we try all combinations within a given size, out of $15$ total augmented prompts ($5$ prompts for each of the three prompt-augmenting models). Figure \ref{fig:CAIRO_results} compares the correlation between fairness metrics resulting from  CAIRO (that uses the best combination of prompts) to the average correlation using all the possible combinations of the prompts. As discussed in Section \ref{sec:why_they_dont_correlate}, unifying the subgroups targeted by fairness metrics leads to higher correlation.

% We also include a baseline that shows the average correlation across all the possible combinations of prompts. A comparison between the average metric correlation using CAIRO and using  

% The correlation of the different fairness metrics for CAIRO and the average correlation are presented in Figure \ref{fig:CAIRO_results}.}

We observe that CAIRO significantly improves the metrics correlation compared to using the original prompts (\textit{i.e.} the number of prompts equals 1). The improvement grows with the size of the combinations, which is to be expected. However, this is not the case for the average baseline, which suggests that simply using all available prompt combinations is not a viable alternative. This showcases the importance of selecting specific prompt combinations to uncover matching patterns across different metrics, as performed by our approach.

% \subsection{How does the correlation between pre-training data distribution and prompt data distribution affect bias?}
% \subsection{Does data augmentation mitigate this issue?}
%\subsection{What are the contributions of the paraphrasing models to CAIRO prompts?}

\begin{figure*}[t]
    \centering
    \includegraphics[trim={0 0cm 0 0},clip,width=0.7\textwidth]{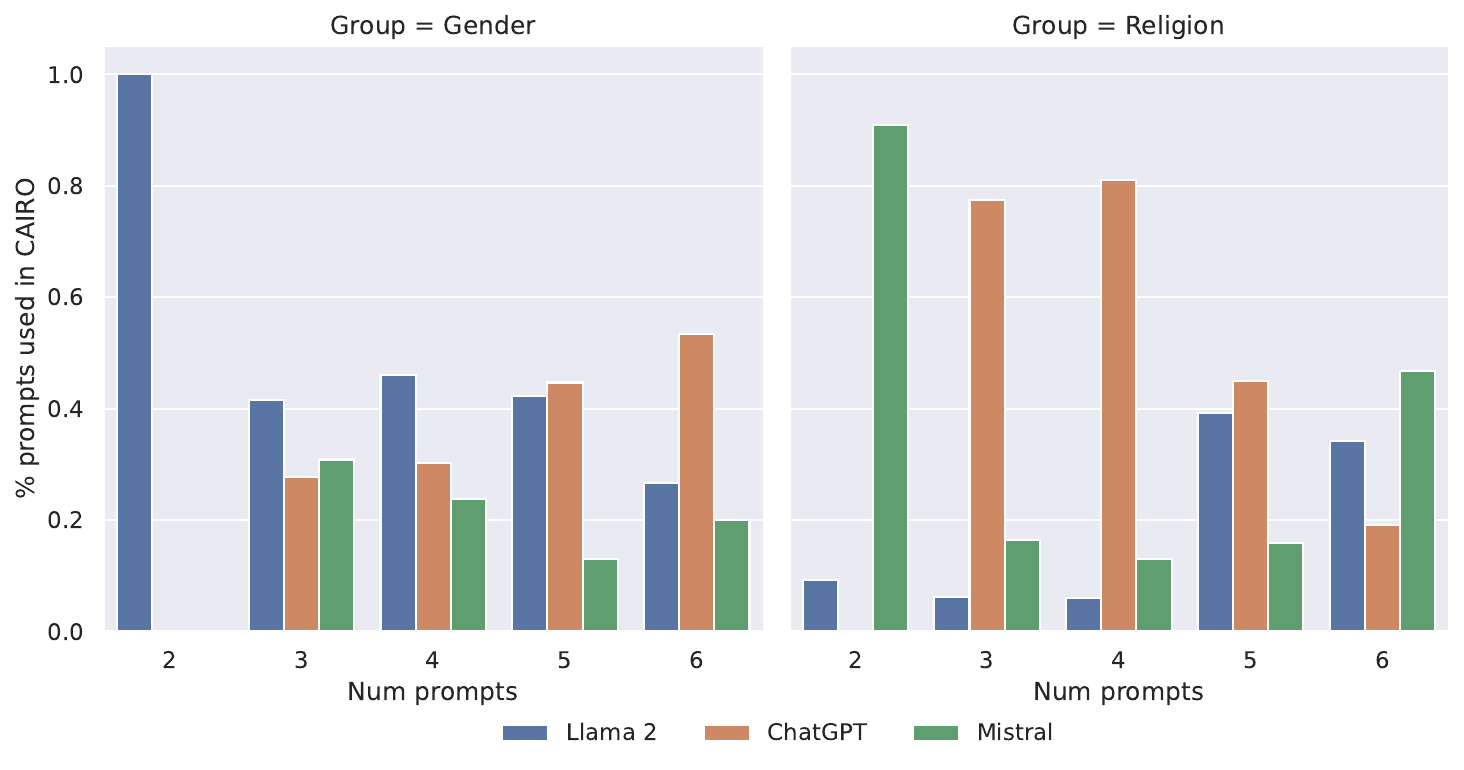}
    \caption{The contributions of the models used to generate the paraphrased prompts with the highest correlation found by CAIRO. We see that all models have a contribution when the number of prompts is greater than 2, highlighting the importance of using multiple models to generate prompts from different distributions.}% As we can observe, CAIRO requires using prompts with different distributions to reduce the effect of the overlap between prompts and the pre-training corpus of the model whose fairness is being assessed.}
    \label{fig:prompt_contributions}
\end{figure*}

\begin{figure*}[t]
\centering   
    \includegraphics[trim={0 0cm 0 0},clip,width=0.7\textwidth]{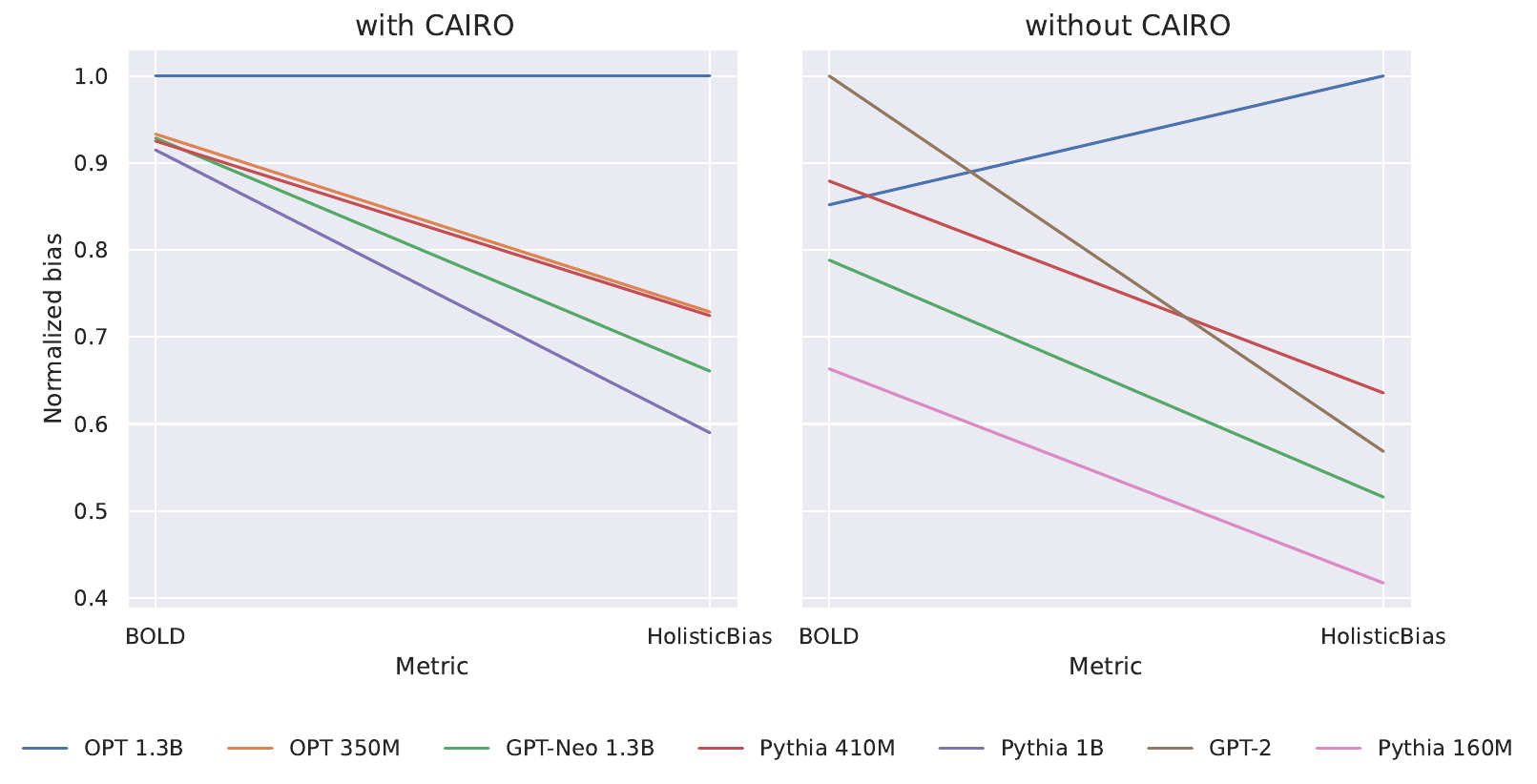}
    \caption{The religion bias values of the top $5$ most biased models (among the list of $10$ models mentioned in Section \ref{sec:experimental_methodology}) according to BOLD and HolisticBias before and after using CAIRO. Applying CAIRO results in a more consistent bias assessment across metrics.}
    \label{fig:ranking}
\end{figure*}

\subsection{What are the contributions of the paraphrasing models to the highest correlated combinations?}\label{sec:prompt_models}

In this experiment, we assess the contributions of each prompt-augmenting model in the combinations that achieved the highest correlation across metrics. The goal of this study is to analyze the importance of having multiple models generating the paraphrases. Results are presented in Figure \ref{fig:prompt_contributions}.
All models contribute to finding the best prompt combination in terms of correlation. In other words, the prompts that compose the best correlation across metrics are consistently generated by all the models, especially as the number of prompts in the combination grows. The only exceptions are observed with a small number of prompts, but this is likely due to the small sample size.

\subsection{How does bias assessment change when using CAIRO?}\label{sec:before_and_after_cairo}

In this final experiment, we study the agreement of the rankings of the models in terms of bias when using the different metrics. In particular, we are interested in analyzing how the original rankings of models that are assessed change after applying CAIRO. The normalized bias of the $5$ most biased models is shown in Figure \ref{fig:ranking}.
The agreement between BOLD and HolisticBias with CAIRO improves compared to without CAIRO. Specifically, both metrics assign the same model as the most biased (OPT $1.3$B) when using CAIRO. However, without CAIRO, the most biased model according to BOLD does not match HolisticBias's. Furthermore, there is a noticeable change in the model rankings in terms of bias across the different metrics without CAIRO. Interestingly, the models with the top-$5$ worst bias change when using CAIRO, with only two models appearing in both scenarios.

\section{Discussion}
{\color{black} The importance of having correlated fairness measurements stems from metrics being only proxies for the bias learned by the model, and increasing the correlation between the metrics could be seen as a signal that we are measuring relevant proxies. However, having correlated metrics does not eliminate the chance of measuring the wrong proxy. Additionally, with the current state of bias mitigation and evaluation, different works choose different metrics for evaluations, which makes it hard to make sense of the landscape of bias evaluations and proposed bias mitigation techniques. Therefore, having correlated metrics is important as it increases the consistency between the measured values by all metrics (as explained in Section \ref{sec:before_and_after_cairo}). If fairness metrics are uncorrelated, an improvement in fairness using one metric will not necessarily lead to an improvement using other metrics (it could lead to fairness degradation on other metrics in the case of negative correlation).}

% {\color{black} We have two main reasons to focus on studying the correlation between prompt-based fairness metrics. First, as mentioned in Section $1$, there exist other works studying the correlation between probability-based and embedding-based fairness metrics [1-2] (although they have not provided a solution to their lack of correlation yet). Second, prompt-based fairness metrics are more recent and have become widely used in the last three years, especially after many papers have come out criticizing the probability-based and embedding-based metrics for not reflecting the bias in the downstream task [3-9], as mentioned in Section 1 L52-58.}
\section{Conclusion}
In this paper, we show that existing prompt-based fairness metrics lack correlation. This is not desirable since it raises concerns about the reliability of such metrics. Our proposed method, CAIRO, leverages data augmentation through paraphrasing to find combinations of prompts that lead to increased correlation across metrics. Ultimately, CAIRO provides a way to reconcile different metrics for a more reliable fairness assessment.

\section*{Acknowledgements} 

{\color{black}{We are thankful to Su Lin Blodgett for her insightful suggestions in this project. We are also thankful to the reviewers for their constructive comments. Sarath Chandar is supported by the Canada CIFAR AI Chairs program, the Canada Research Chair in Lifelong Machine Learning, and the NSERC Discovery Grant. Abdelrahman Zayed is supported by an FRQNT scholarship. Gonçalo Mordido was supported by an FRQNT postdoctoral scholarship (PBEEE) during part of this work. The authors acknowledge the computational resources provided by the Digital Research Alliance of Canada and Microsoft Research.}}
\section*{Limitations and Ethical Considerations}\label{app:limitations}
Our work aims to enhance the reliability of fairness assessment across various prompt-based metrics. However, it relies on the assumption that these metrics target similar or overlapping demographic subgroups. For instance, if one metric focuses on race bias with Black and White subgroups, while another metric targets Chinese and Arab subgroups, applying our method, CAIRO, may not necessarily enhance their correlation. Another limitation arises from the similarity of lexical semantics in the bias metrics used. Substantial differences in lexical semantics could result in a lack of correlation between metric values even after applying CAIRO. 

Additionally, CAIRO assumes that the prompts used for data augmentation originate from distinct distributions, as they are generated by models trained on different corpora (ChatGPT, Llama 2, and Mistral). However, if paraphrasing models have significant overlap in their training data, the improvement in metric correlation using CAIRO may be less pronounced. We also acknowledge that CAIRO can be used in an alternative way to search for prompts that maximize other criteria such as toxic output.
\bibliography{custom}

\begin{thebibliography}{55}
\expandafter\ifx\csname natexlab\endcsname\relax\def\natexlab#1{#1}\fi

\bibitem[{Attanasio et~al.(2022)Attanasio, Nozza, Hovy, and Baralis}]{attanasio2022entropy}
Giuseppe Attanasio, Debora Nozza, Dirk Hovy, and Elena Baralis. 2022.
\newblock Entropy-based attention regularization frees unintended bias mitigation from lists.
\newblock In \emph{Findings of the Association for Computational Linguistics: ACL 2022}. Association for Computational Linguistics.

\bibitem[{Bassignana et~al.(2018)Bassignana, Basile, Patti et~al.}]{bassignana2018hurtlex}
Elisa Bassignana, Valerio Basile, Viviana Patti, et~al. 2018.
\newblock Hurtlex: A multilingual lexicon of words to hurt.
\newblock In \emph{CEUR Workshop proceedings}, volume 2253, pages 1--6. CEUR-WS.

\bibitem[{Bhagat and Hovy(2013)}]{10.1162/COLI_a_00166}
Rahul Bhagat and Eduard Hovy. 2013.
\newblock \href {https://doi.org/10.1162/COLI_a_00166} {{What Is a Paraphrase?}}
\newblock \emph{Computational Linguistics}, 39(3):463--472.

\bibitem[{Biderman et~al.(2023)Biderman, Schoelkopf, Anthony, Bradley, O’Brien, Hallahan, Khan, Purohit, Prashanth, Raff et~al.}]{biderman2023pythia}
Stella Biderman, Hailey Schoelkopf, Quentin~Gregory Anthony, Herbie Bradley, Kyle O’Brien, Eric Hallahan, Mohammad~Aflah Khan, Shivanshu Purohit, USVSN~Sai Prashanth, Edward Raff, et~al. 2023.
\newblock Pythia: A suite for analyzing large language models across training and scaling.
\newblock In \emph{International Conference on Machine Learning}, pages 2397--2430. PMLR.

\bibitem[{Black et~al.(2021)Black, Leo, Wang, Leahy, and Biderman}]{gpt-neo}
Sid Black, Gao Leo, Phil Wang, Connor Leahy, and Stella Biderman. 2021.
\newblock \href {https://doi.org/10.5281/zenodo.5297715} {{GPT-Neo: Large Scale Autoregressive Language Modeling with Mesh-Tensorflow}}.
\newblock {If you use this software, please cite it using these metadata.}

\bibitem[{Blodgett et~al.(2021)Blodgett, Lopez, Olteanu, Sim, and Wallach}]{blodgett2021stereotyping}
Su~Lin Blodgett, Gilsinia Lopez, Alexandra Olteanu, Robert Sim, and Hanna Wallach. 2021.
\newblock Stereotyping norwegian salmon: An inventory of pitfalls in fairness benchmark datasets.
\newblock In \emph{Proceedings of the 59th Annual Meeting of the Association for Computational Linguistics and the 11th International Joint Conference on Natural Language Processing (Volume 1: Long Papers)}, pages 1004--1015.

\bibitem[{Bordia and Bowman(2019)}]{bordia-bowman-2019-identifying}
Shikha Bordia and Samuel~R. Bowman. 2019.
\newblock \href {https://doi.org/10.18653/v1/N19-3002} {Identifying and reducing gender bias in word-level language models}.
\newblock In \emph{Proceedings of the 2019 Conference of the North {A}merican Chapter of the Association for Computational Linguistics: Student Research Workshop}, pages 7--15, Minneapolis, Minnesota. Association for Computational Linguistics.

\bibitem[{Cabello et~al.(2023)Cabello, J{\o}rgensen, and S{\o}gaard}]{cabello2023independence}
Laura Cabello, Anna~Katrine J{\o}rgensen, and Anders S{\o}gaard. 2023.
\newblock On the independence of association bias and empirical fairness in language models.
\newblock In \emph{Proceedings of the 2023 ACM Conference on Fairness, Accountability, and Transparency}, pages 370--378.

\bibitem[{Caliskan et~al.(2017)Caliskan, Bryson, and Narayanan}]{caliskan2017semantics}
Aylin Caliskan, Joanna~J Bryson, and Arvind Narayanan. 2017.
\newblock Semantics derived automatically from language corpora contain human-like biases.
\newblock \emph{Science}, 356(6334):183--186.

\bibitem[{Cao et~al.(2022{\natexlab{a}})Cao, Lin, Han, Liu, and Sun}]{cao-etal-2022-prompt}
Boxi Cao, Hongyu Lin, Xianpei Han, Fangchao Liu, and Le~Sun. 2022{\natexlab{a}}.
\newblock \href {https://doi.org/10.18653/v1/2022.acl-long.398} {Can prompt probe pretrained language models? understanding the invisible risks from a causal view}.
\newblock In \emph{Proceedings of the 60th Annual Meeting of the Association for Computational Linguistics (Volume 1: Long Papers)}, pages 5796--5808, Dublin, Ireland. Association for Computational Linguistics.

\bibitem[{Cao et~al.(2021)Cao, Lin, Han, Sun, Yan, Liao, Xue, and Xu}]{cao-etal-2021-knowledgeable}
Boxi Cao, Hongyu Lin, Xianpei Han, Le~Sun, Lingyong Yan, Meng Liao, Tong Xue, and Jin Xu. 2021.
\newblock \href {https://doi.org/10.18653/v1/2021.acl-long.146} {Knowledgeable or educated guess? revisiting language models as knowledge bases}.
\newblock In \emph{Proceedings of the 59th Annual Meeting of the Association for Computational Linguistics and the 11th International Joint Conference on Natural Language Processing (Volume 1: Long Papers)}, pages 1860--1874, Online. Association for Computational Linguistics.

\bibitem[{Cao et~al.(2022{\natexlab{b}})Cao, Pruksachatkun, Chang, Gupta, Kumar, Dhamala, and Galstyan}]{cao-etal-2022-intrinsic}
Yang~Trista Cao, Yada Pruksachatkun, Kai-Wei Chang, Rahul Gupta, Varun Kumar, Jwala Dhamala, and Aram Galstyan. 2022{\natexlab{b}}.
\newblock \href {https://doi.org/10.18653/v1/2022.acl-short.62} {On the intrinsic and extrinsic fairness evaluation metrics for contextualized language representations}.
\newblock In \emph{Proceedings of the 60th Annual Meeting of the Association for Computational Linguistics (Volume 2: Short Papers)}, pages 561--570, Dublin, Ireland. Association for Computational Linguistics.

\bibitem[{Delobelle et~al.(2022)Delobelle, Tokpo, Calders, and Berendt}]{delobelle-etal-2022-measuring}
Pieter Delobelle, Ewoenam Tokpo, Toon Calders, and Bettina Berendt. 2022.
\newblock \href {https://doi.org/10.18653/v1/2022.naacl-main.122} {Measuring fairness with biased rulers: A comparative study on bias metrics for pre-trained language models}.
\newblock In \emph{Proceedings of the 2022 Conference of the North American Chapter of the Association for Computational Linguistics: Human Language Technologies}, pages 1693--1706, Seattle, United States. Association for Computational Linguistics.

\bibitem[{Dhamala et~al.(2021)Dhamala, Sun, Kumar, Krishna, Pruksachatkun, Chang, and Gupta}]{dhamala2021bold}
Jwala Dhamala, Tony Sun, Varun Kumar, Satyapriya Krishna, Yada Pruksachatkun, Kai-Wei Chang, and Rahul Gupta. 2021.
\newblock Bold: Dataset and metrics for measuring biases in open-ended language generation.
\newblock In \emph{Proceedings of the 2021 ACM conference on fairness, accountability, and transparency}, pages 862--872.

\bibitem[{Dixon et~al.(2018)Dixon, Li, Sorensen, Thain, and Vasserman}]{dixon2018measuring}
Lucas Dixon, John Li, Jeffrey Sorensen, Nithum Thain, and Lucy Vasserman. 2018.
\newblock Measuring and mitigating unintended bias in text classification.
\newblock In \emph{Conference on AI, Ethics, and Society}.

\bibitem[{Elazar et~al.(2021)Elazar, Kassner, Ravfogel, Ravichander, Hovy, Sch{\"u}tze, and Goldberg}]{elazar2021measuring}
Yanai Elazar, Nora Kassner, Shauli Ravfogel, Abhilasha Ravichander, Eduard Hovy, Hinrich Sch{\"u}tze, and Yoav Goldberg. 2021.
\newblock Measuring and improving consistency in pretrained language models.
\newblock \emph{Transactions of the Association for Computational Linguistics}, 9:1012--1031.

\bibitem[{Gallegos et~al.(2023)Gallegos, Rossi, Barrow, Tanjim, Kim, Dernoncourt, Yu, Zhang, and Ahmed}]{gallegos2023bias}
Isabel~O Gallegos, Ryan~A Rossi, Joe Barrow, Md~Mehrab Tanjim, Sungchul Kim, Franck Dernoncourt, Tong Yu, Ruiyi Zhang, and Nesreen~K Ahmed. 2023.
\newblock Bias and fairness in large language models: A survey.
\newblock \emph{arXiv preprint arXiv:2309.00770}.

\bibitem[{Goldfarb-Tarrant et~al.(2021)Goldfarb-Tarrant, Marchant, Mu{\~n}oz~S{\'a}nchez, Pandya, and Lopez}]{goldfarb-tarrant-etal-2021-intrinsic}
Seraphina Goldfarb-Tarrant, Rebecca Marchant, Ricardo Mu{\~n}oz~S{\'a}nchez, Mugdha Pandya, and Adam Lopez. 2021.
\newblock \href {https://doi.org/10.18653/v1/2021.acl-long.150} {Intrinsic bias metrics do not correlate with application bias}.
\newblock In \emph{Proceedings of the 59th Annual Meeting of the Association for Computational Linguistics and the 11th International Joint Conference on Natural Language Processing (Volume 1: Long Papers)}, pages 1926--1940, Online. Association for Computational Linguistics.

\bibitem[{Huang et~al.(2020)Huang, Zhang, Jiang, Stanforth, Welbl, Rae, Maini, Yogatama, and Kohli}]{huang-etal-2020-reducing}
Po-Sen Huang, Huan Zhang, Ray Jiang, Robert Stanforth, Johannes Welbl, Jack Rae, Vishal Maini, Dani Yogatama, and Pushmeet Kohli. 2020.
\newblock \href {https://doi.org/10.18653/v1/2020.findings-emnlp.7} {Reducing sentiment bias in language models via counterfactual evaluation}.
\newblock In \emph{Findings of the Association for Computational Linguistics: EMNLP 2020}, pages 65--83, Online. Association for Computational Linguistics.

\bibitem[{Jiang et~al.(2023)Jiang, Sablayrolles, Mensch, Bamford, Chaplot, Casas, Bressand, Lengyel, Lample, Saulnier et~al.}]{jiang2023mistral}
Albert~Q Jiang, Alexandre Sablayrolles, Arthur Mensch, Chris Bamford, Devendra~Singh Chaplot, Diego de~las Casas, Florian Bressand, Gianna Lengyel, Guillaume Lample, Lucile Saulnier, et~al. 2023.
\newblock Mistral 7b.
\newblock \emph{arXiv preprint arXiv:2310.06825}.

\bibitem[{Kaneko et~al.(2022)Kaneko, Bollegala, and Okazaki}]{kaneko-etal-2022-debiasing}
Masahiro Kaneko, Danushka Bollegala, and Naoaki Okazaki. 2022.
\newblock \href {https://aclanthology.org/2022.coling-1.111} {Debiasing isn{'}t enough! {--} on the effectiveness of debiasing {MLM}s and their social biases in downstream tasks}.
\newblock In \emph{Proceedings of the 29th International Conference on Computational Linguistics}, pages 1299--1310, Gyeongju, Republic of Korea. International Committee on Computational Linguistics.

\bibitem[{Kurita et~al.(2019{\natexlab{a}})Kurita, Vyas, Pareek, Black, and Tsvetkov}]{kurita2019measuring}
Keita Kurita, Nidhi Vyas, Ayush Pareek, Alan~W Black, and Yulia Tsvetkov. 2019{\natexlab{a}}.
\newblock Measuring bias in contextualized word representations.
\newblock In \emph{Proceedings of the First Workshop on Gender Bias in Natural Language Processing}, pages 166--172.

\bibitem[{Kurita et~al.(2019{\natexlab{b}})Kurita, Vyas, Pareek, Black, and Tsvetkov}]{kurita-etal-2019-measuring}
Keita Kurita, Nidhi Vyas, Ayush Pareek, Alan~W Black, and Yulia Tsvetkov. 2019{\natexlab{b}}.
\newblock \href {https://doi.org/10.18653/v1/W19-3823} {Measuring bias in contextualized word representations}.
\newblock In \emph{Proceedings of the First Workshop on Gender Bias in Natural Language Processing}, pages 166--172, Florence, Italy. Association for Computational Linguistics.

\bibitem[{Li et~al.(2020{\natexlab{a}})Li, Feng, Meng, Han, Wu, and Li}]{li2019unified}
Xiaoya Li, Jingrong Feng, Yuxian Meng, Qinghong Han, Fei Wu, and Jiwei Li. 2020{\natexlab{a}}.
\newblock A unified {MRC} framework for named entity recognition.
\newblock In \emph{Proceedings of the 58th Annual Meeting of the Association for Computational Linguistics}, pages 5849--5859, Online. Association for Computational Linguistics.

\bibitem[{Li et~al.(2020{\natexlab{b}})Li, Sun, Meng, Liang, Wu, and Li}]{li2019dice}
Xiaoya Li, Xiaofei Sun, Yuxian Meng, Junjun Liang, Fei Wu, and Jiwei Li. 2020{\natexlab{b}}.
\newblock Dice loss for data-imbalanced {NLP} tasks.
\newblock In \emph{Proceedings of the 58th Annual Meeting of the Association for Computational Linguistics}, pages 465--476, Online. Association for Computational Linguistics.

\bibitem[{Liu et~al.(2023)Liu, Zheng, Du, Ding, Qian, Yang, and Tang}]{LIU2023}
Xiao Liu, Yanan Zheng, Zhengxiao Du, Ming Ding, Yujie Qian, Zhilin Yang, and Jie Tang. 2023.
\newblock \href {https://doi.org/https://doi.org/10.1016/j.aiopen.2023.08.012} {Gpt understands, too}.
\newblock \emph{AI Open}.

\bibitem[{Liu et~al.(2022)Liu, Liu, Radev, and Neubig}]{liu2022brio}
Yixin Liu, Pengfei Liu, Dragomir Radev, and Graham Neubig. 2022.
\newblock {BRIO}: Bringing order to abstractive summarization.
\newblock In \emph{Proceedings of the 60th Annual Meeting of the Association for Computational Linguistics (Volume 1: Long Papers)}, pages 2890--2903, Dublin, Ireland. Association for Computational Linguistics.

\bibitem[{Lu et~al.(2020)Lu, Mardziel, Wu, Amancharla, and Datta}]{lu2020gender}
Kaiji Lu, Piotr Mardziel, Fangjing Wu, Preetam Amancharla, and Anupam Datta. 2020.
\newblock Gender bias in neural natural language processing.
\newblock In \emph{Logic, Language, and Security}.

\bibitem[{May et~al.(2019)May, Wang, Bordia, Bowman, and Rudinger}]{may2019measuring}
Chandler May, Alex Wang, Shikha Bordia, Samuel~R. Bowman, and Rachel Rudinger. 2019.
\newblock On measuring social biases in sentence encoders.
\newblock In \emph{Conference of the North {A}merican Chapter of the Association for Computational Linguistics}.

\bibitem[{Mei et~al.(2023)Mei, Fereidooni, and Caliskan}]{mei2023bias}
Katelyn Mei, Sonia Fereidooni, and Aylin Caliskan. 2023.
\newblock Bias against 93 stigmatized groups in masked language models and downstream sentiment classification tasks.
\newblock In \emph{Proceedings of the 2023 ACM Conference on Fairness, Accountability, and Transparency}, pages 1699--1710.

\bibitem[{Mozafari et~al.(2020)Mozafari, Farahbakhsh, and Crespi}]{mozafari2020hate}
Marzieh Mozafari, Reza Farahbakhsh, and No{\"e}l Crespi. 2020.
\newblock Hate speech detection and racial bias mitigation in social media based on bert model.
\newblock \emph{PloS one}, 15(8):e0237861.

\bibitem[{Nadeem et~al.(2021)Nadeem, Bethke, and Reddy}]{nadeem2021stereoset}
Moin Nadeem, Anna Bethke, and Siva Reddy. 2021.
\newblock Stereoset: Measuring stereotypical bias in pretrained language models.
\newblock In \emph{Proceedings of the 59th Annual Meeting of the Association for Computational Linguistics and the 11th International Joint Conference on Natural Language Processing (Volume 1: Long Papers)}, pages 5356--5371.

\bibitem[{Nangia et~al.(2020)Nangia, Vania, Bhalerao, and Bowman}]{nangia-etal-2020-crows}
Nikita Nangia, Clara Vania, Rasika Bhalerao, and Samuel~R. Bowman. 2020.
\newblock \href {https://doi.org/10.18653/v1/2020.emnlp-main.154} {{C}row{S}-pairs: A challenge dataset for measuring social biases in masked language models}.
\newblock In \emph{Proceedings of the 2020 Conference on Empirical Methods in Natural Language Processing (EMNLP)}, pages 1953--1967, Online. Association for Computational Linguistics.

\bibitem[{Nozza et~al.(2021)Nozza, Bianchi, Hovy et~al.}]{nozza2021honest}
Debora Nozza, Federico Bianchi, Dirk Hovy, et~al. 2021.
\newblock Honest: Measuring hurtful sentence completion in language models.
\newblock In \emph{Proceedings of the 2021 Conference of the North American Chapter of the Association for Computational Linguistics: Human Language Technologies}. Association for Computational Linguistics.

\bibitem[{Orgad and Belinkov(2023)}]{orgad-belinkov-2023-blind}
Hadas Orgad and Yonatan Belinkov. 2023.
\newblock \href {https://doi.org/10.18653/v1/2023.acl-long.490} {{BLIND}: Bias removal with no demographics}.
\newblock In \emph{Proceedings of the 61st Annual Meeting of the Association for Computational Linguistics (Volume 1: Long Papers)}, pages 8801--8821, Toronto, Canada. Association for Computational Linguistics.

\bibitem[{Orgad et~al.(2022)Orgad, Goldfarb-Tarrant, and Belinkov}]{orgad-etal-2022-gender}
Hadas Orgad, Seraphina Goldfarb-Tarrant, and Yonatan Belinkov. 2022.
\newblock \href {https://doi.org/10.18653/v1/2022.naacl-main.188} {How gender debiasing affects internal model representations, and why it matters}.
\newblock In \emph{Proceedings of the 2022 Conference of the North American Chapter of the Association for Computational Linguistics: Human Language Technologies}, pages 2602--2628, Seattle, United States. Association for Computational Linguistics.

\bibitem[{Petroni et~al.(2019)Petroni, Rockt{\"a}schel, Riedel, Lewis, Bakhtin, Wu, and Miller}]{petroni-etal-2019-language}
Fabio Petroni, Tim Rockt{\"a}schel, Sebastian Riedel, Patrick Lewis, Anton Bakhtin, Yuxiang Wu, and Alexander Miller. 2019.
\newblock \href {https://doi.org/10.18653/v1/D19-1250} {Language models as knowledge bases?}
\newblock In \emph{Proceedings of the 2019 Conference on Empirical Methods in Natural Language Processing and the 9th International Joint Conference on Natural Language Processing (EMNLP-IJCNLP)}, pages 2463--2473, Hong Kong, China. Association for Computational Linguistics.

\bibitem[{Poerner et~al.(2020)Poerner, Waltinger, and Sch{\"u}tze}]{poerner-etal-2020-e}
Nina Poerner, Ulli Waltinger, and Hinrich Sch{\"u}tze. 2020.
\newblock \href {https://doi.org/10.18653/v1/2020.findings-emnlp.71} {{E}-{BERT}: Efficient-yet-effective entity embeddings for {BERT}}.
\newblock In \emph{Findings of the Association for Computational Linguistics: EMNLP 2020}, pages 803--818, Online. Association for Computational Linguistics.

\bibitem[{Radford et~al.(2019)Radford, Wu, Child, Luan, Amodei, and Sutskever}]{radford2019language}
Alec Radford, Jeffrey Wu, Rewon Child, David Luan, Dario Amodei, and Ilya Sutskever. 2019.
\newblock {Language Models are Unsupervised Multitask Learners}.
\newblock \emph{OpenAI Blog}, 1(8):9.

\bibitem[{Rajpurkar et~al.(2018)Rajpurkar, Jia, and Liang}]{rajpurkar2018know}
Pranav Rajpurkar, Robin Jia, and Percy Liang. 2018.
\newblock Know what you don{'}t know: Unanswerable questions for {SQ}u{AD}.
\newblock In \emph{Proceedings of the 56th Annual Meeting of the Association for Computational Linguistics (Volume 2: Short Papers)}, pages 784--789, Melbourne, Australia. Association for Computational Linguistics.

\bibitem[{Rajpurkar et~al.(2016)Rajpurkar, Zhang, Lopyrev, and Liang}]{rajpurkar2016squad}
Pranav Rajpurkar, Jian Zhang, Konstantin Lopyrev, and Percy Liang. 2016.
\newblock {SQ}u{AD}: 100,000+ questions for machine comprehension of text.
\newblock In \emph{Conference on Empirical Methods in Natural Language Processing}.

\bibitem[{Sap et~al.(2019)Sap, Card, Gabriel, Choi, and Smith}]{sap2019risk}
Maarten Sap, Dallas Card, Saadia Gabriel, Yejin Choi, and Noah~A Smith. 2019.
\newblock The risk of racial bias in hate speech detection.
\newblock In \emph{Proceedings of the 57th annual meeting of the association for computational linguistics}, pages 1668--1678.

\bibitem[{Sicilia and Alikhani(2023)}]{sicilia-alikhani-2023-learning}
Anthony Sicilia and Malihe Alikhani. 2023.
\newblock \href {https://doi.org/10.18653/v1/2023.acl-long.163} {Learning to generate equitable text in dialogue from biased training data}.
\newblock In \emph{Proceedings of the 61st Annual Meeting of the Association for Computational Linguistics (Volume 1: Long Papers)}, pages 2898--2917, Toronto, Canada. Association for Computational Linguistics.

\bibitem[{Smith et~al.(2022)Smith, Hall, Kambadur, Presani, and Williams}]{smith2022m}
Eric~Michael Smith, Melissa Hall, Melanie Kambadur, Eleonora Presani, and Adina Williams. 2022.
\newblock “i’m sorry to hear that”: Finding new biases in language models with a holistic descriptor dataset.
\newblock In \emph{Proceedings of the 2022 Conference on Empirical Methods in Natural Language Processing}, pages 9180--9211.

\bibitem[{Steed et~al.(2022)Steed, Panda, Kobren, and Wick}]{steed-etal-2022-upstream}
Ryan Steed, Swetasudha Panda, Ari Kobren, and Michael Wick. 2022.
\newblock \href {https://doi.org/10.18653/v1/2022.acl-long.247} {{U}pstream {M}itigation {I}s \textit{ {N}ot} {A}ll {Y}ou {N}eed: {T}esting the {B}ias {T}ransfer {H}ypothesis in {P}re-{T}rained {L}anguage {M}odels}.
\newblock In \emph{Proceedings of the 60th Annual Meeting of the Association for Computational Linguistics (Volume 1: Long Papers)}, pages 3524--3542, Dublin, Ireland. Association for Computational Linguistics.

\bibitem[{Touvron et~al.(2023)Touvron, Martin, Stone, Albert, Almahairi, Babaei, Bashlykov, Batra, Bhargava, Bhosale et~al.}]{touvron2023llama}
Hugo Touvron, Louis Martin, Kevin Stone, Peter Albert, Amjad Almahairi, Yasmine Babaei, Nikolay Bashlykov, Soumya Batra, Prajjwal Bhargava, Shruti Bhosale, et~al. 2023.
\newblock Llama 2: Open foundation and fine-tuned chat models.
\newblock \emph{arXiv preprint arXiv:2307.09288}.

\bibitem[{Vaswani et~al.(2017)Vaswani, Shazeer, Parmar, Uszkoreit, Jones, Gomez, Kaiser, and Polosukhin}]{vaswani2017attention}
Ashish Vaswani, Noam Shazeer, Niki Parmar, Jakob Uszkoreit, Llion Jones, Aidan~N Gomez, {\L}ukasz Kaiser, and Illia Polosukhin. 2017.
\newblock {Attention is All You Need}.
\newblock In \emph{{Proc. of the Advances in Neural Information Processing Systems (Neurips)}}, pages 5998--6008.

\bibitem[{Wang et~al.(2018)Wang, Singh, Michael, Hill, Levy, and Bowman}]{wang2018glue}
Alex Wang, Amanpreet Singh, Julian Michael, Felix Hill, Omer Levy, and Samuel Bowman. 2018.
\newblock {GLUE}: A multi-task benchmark and analysis platform for natural language understanding.
\newblock In \emph{{EMNLP} Workshop {B}lackbox{NLP}: Analyzing and Interpreting Neural Networks for {NLP}}.

\bibitem[{Wang and Komatsuzaki(2021)}]{gpt-j}
Ben Wang and Aran Komatsuzaki. 2021.
\newblock {GPT-J-6B: A 6 Billion Parameter Autoregressive Language Model}.
\newblock \url{https://github.com/kingoflolz/mesh-transformer-jax}.

\bibitem[{Webster et~al.(2020)Webster, Wang, Tenney, Beutel, Pitler, Pavlick, Chen, Chi, and Petrov}]{webster2020measuring}
Kellie Webster, Xuezhi Wang, Ian Tenney, Alex Beutel, Emily Pitler, Ellie Pavlick, Jilin Chen, Ed~H Chi, and Slav Petrov. 2020.
\newblock Measuring and reducing gendered correlations in pre-trained models.

\bibitem[{Yu et~al.(2020)Yu, Bohnet, and Poesio}]{yu-etal-2020-named}
Juntao Yu, Bernd Bohnet, and Massimo Poesio. 2020.
\newblock \href {https://doi.org/10.18653/v1/2020.acl-main.577} {Named entity recognition as dependency parsing}.
\newblock In \emph{Proceedings of the 58th Annual Meeting of the Association for Computational Linguistics}, pages 6470--6476, Online. Association for Computational Linguistics.

\bibitem[{Zayed et~al.(2024)Zayed, Mordido, Shabanian, Baldini, and Chandar}]{zayed2023fairnessaware}
Abdelrahman Zayed, Goncalo Mordido, Samira Shabanian, Ioana Baldini, and Sarath Chandar. 2024.
\newblock Fairness-aware structured pruning in transformers.
\newblock In \emph{AAAI Conference on Artificial Intelligence}.

\bibitem[{Zayed et~al.(2023)Zayed, Parthasarathi, Mordido, Palangi, Shabanian, and Chandar}]{zayed2022deep}
Abdelrahman Zayed, Prasanna Parthasarathi, Goncalo Mordido, Hamid Palangi, Samira Shabanian, and Sarath Chandar. 2023.
\newblock Deep learning on a healthy data diet: Finding important examples for fairness.
\newblock In \emph{AAAI Conference on Artificial Intelligence}.

\bibitem[{Zhang et~al.(2022)Zhang, Roller, Goyal, Artetxe, Chen, Chen, Dewan, Diab, Li, Lin, Mihaylov, Ott, Shleifer, Shuster, Simig, Koura, Sridhar, Wang, and Zettlemoyer}]{zhang2022opt}
Susan Zhang, Stephen Roller, Naman Goyal, Mikel Artetxe, Moya Chen, Shuohui Chen, Christopher Dewan, Mona Diab, Xian Li, Xi~Victoria Lin, Todor Mihaylov, Myle Ott, Sam Shleifer, Kurt Shuster, Daniel Simig, Punit~Singh Koura, Anjali Sridhar, Tianlu Wang, and Luke Zettlemoyer. 2022.
\newblock \href {http://arxiv.org/abs/2205.01068} {Opt: Open pre-trained transformer language models}.

\bibitem[{Zhang et~al.(2020)Zhang, Zhou, and Li}]{ijcai2020p560}
Yu~Zhang, Houquan Zhou, and Zhenghua Li. 2020.
\newblock \href {https://doi.org/10.24963/ijcai.2020/560} {Fast and accurate neural crf constituency parsing}.
\newblock In \emph{Proceedings of the Twenty-Ninth International Joint Conference on Artificial Intelligence, {IJCAI-20}}, pages 4046--4053. International Joint Conferences on Artificial Intelligence Organization.
\newblock Main track.

\end{thebibliography}

\appendix

\section{Implementation details}
\label{app:imp_details}

{\color{black}This section provides the implementation details regarding running time, the infrastructure used, text generation configurations, and paraphrasing prompts.}
% Our work can be used to increase the reliability of fairness assessment between prompt-based metrics. However, it assumes that the used metrics measure the bias targeting the same or overlapping subgroups. For example, if one metric considers the targeted subgroups from race bias to be Black and White, and another metric considers Chinese and Arab as the targeted subgroups, CAIRO might not lead to an improvement in the correlation between them. Another limitation is on the similarity of the lexical semantics on the bias metrics used. Significant differences in the lexical semantics might lead to decorrelation in the metric values, even after applying CAIRO. Another limitation is that CAIRO assumes that the modes being used for generating the data-augmented prompts come from different distributions since the models that generated them have been trained on different corpora (ChatGPT, Llama $2$, and Mistral). However, using paraphrasing models with significant overlap in their training corpora will lead to less improvement in metric correlation using CAIRO.

% \subsection{Number of trainable parameters}\label{app:impl_num_par}
% \textcolor{black}{In text classification}, our experiments were conducted on BERT \cite{devlin2018bert} and RoBERTa \cite{liu2019roberta} base models, which possess $110$ and $125$ million trainable parameters, respectively. \textcolor{black}{As for text generation, we used GPT-Neo \cite{gpt-neo} with $1.3$ and $2.7$ billion parameters.}

\subsection{Infrastructure used}\label{app:impl_infra_str}
{\color{black}We used Tesla P100-PCIE-12GB GPU. The necessary packages to execute the code are included in our code's $requirements.txt$ file.}

\subsection{Running time}\label{app:impl_time}
% The running time for each experiment varied based on the dataset size. Using $1$ GPU, the running time was $4$ hours, $12$ hours, and $24$ hours, for the Twitter, Wikipedia, and Jigsaw datasets, respectively. 

The computational time for each experiment is proportional to the size of the corresponding prompt-based metric. Using a single GPU, the running time was approximately $3$, $6$, and $12$ hours for HONEST, BOLD, and HolisticBias metrics.

\subsection{Decoding configurations for text generation}\label{app:impl_config}
We applied the following configurations:

\begin{itemize}
    \item The maximum allowed tokens for generation, excluding the prompt tokens is $25$ tokens.
    \item The minimum required tokens for generation, without considering the prompt tokens is $0$ tokens.
    \item We employed sampling, instead of using greedy decoding.
    \item No beam search was utilized.
\end{itemize}
{\color{black}{The temperature, top p, and maximum tokens for ChatGPT, Llama 2, and Mistral are $0.95$, $1$, $800$; $0.6$, $0.9$, $4096$; and $0.6$, $0.9$, $4096$, respectively.}}
\subsection{Paraphrasing prompts}
{\color{black}{For ChatGPT, we used the following prompt to get the paraphrases: ``Paraphrase each of the following while not writing the original sentences: [the original prompt]''. For text completion models, namely Llama $2$ and Mistral, we used the following prompt: ``[the original prompt] can be paraphrased as...}''.}

\section{Algorithm used in CAIRO}\label{app:exp_details}

The algorithm used to find the best combination of prompts to maximize the correlation between fairness metrics is described below:
\begin{algorithm}[]
\caption{Correlated Fairness output (CAIRO)}
\textbf{Input:} A set of $A$ language models from $a_{1}$ to $a_{A}$ whose fairness is to be assessed, $M$ metrics from $m_{1}$ to $m_{M}$ used for fairness assessment, $P$ prompt generation language models from $P_{1}$ to $P_{P}$. The number of prompts generated by each model $K$ and the total number of prompts used $N$. The bias quantification $Q$.
\begin{algorithmic}[1]
    \FOR{$metric\in\{m_{1},...,m_{M}\}$}
    % \STATE{
      \STATE{\textrm{ $metric$.$bias\char`_ quantification$ } = $Q$}
      \FOR{$model\in\{P_{1},...,P_{P}\}$} 
      \FOR{$i\in\{1,...,K\}$} 
      \STATE{$metric$.$prompts$$+=$$model.prompt$} 
      \ENDFOR
      % \STATE{\textrm{ $metric$.$prompts.append(prompt)$ }} 
      \ENDFOR
    % } 
    \ENDFOR

        % \STATE{$best\char`_prompts$=[]}
      \FOR{$(metric_1, metric_2)\in\{(m_1,m_2),...\}$} 
      \STATE{$best\char`_prompts$=[]}
      \FOR{$prompt_1\in\{metric_1.prompts\}$} 
      % \STATE{}
      \FOR{$prompt_2\in\{metric_2.prompts\}$}  
      % \STATE{}
      % \STATE{$metric_1.best_prompt$}
      \STATE{$corr(metric_{1},metric_{2})\char`_max=-1$}
      \FOR{$model\in\{A_{1},...,A_{A}\}$} 
       \item $bias_1$($model$)=$metric_{1}$($model$)
       \item $bias_2$($model$)=$metric_{2}$($model$)
      \ENDFOR

    \IF{$corr(metric_{1},metric_{2})$ $>$ $corr(metric_{1},metric_{2})\char`_max$}
        \STATE $corr(metric_{1},metric_{2})\char`_max$\\=$corr(metric_{1},metric_{2})$
        \STATE $prompt^*_1$=$prompt_1$
        \STATE $prompt^*_2$=$prompt_2$
    \ENDIF      
      \ENDFOR
      \ENDFOR      
      \STATE{$best\char`_prompts+=[(prompt^*_1,prompt^*_2)]$}
      \ENDFOR
\end{algorithmic}
\end{algorithm}
\label{alg:cairo}
{\color{black}
\section{Frequently asked questions}\label{sec:FAQ}
This section answers some of the frequently asked questions regarding our work. 

\begin{figure*}[h]
     \centering
     \begin{subfigure}
    \centering    \includegraphics[width=0.49\linewidth]{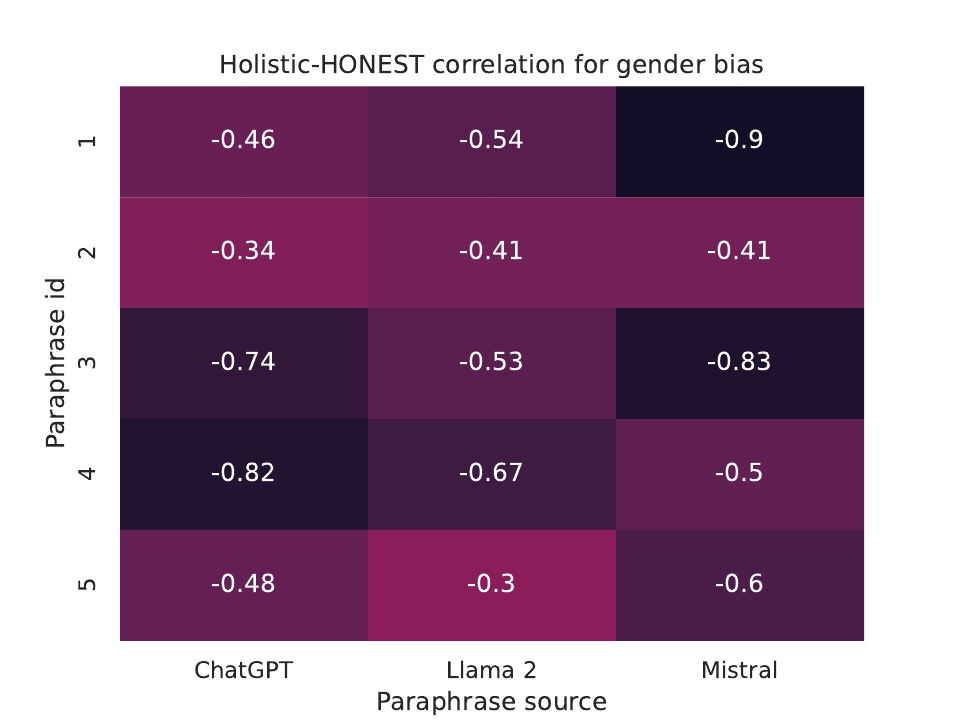}
     %\caption{Fairness}
     \end{subfigure}
     \begin{subfigure}
    \centering    \includegraphics[width=0.49\linewidth]{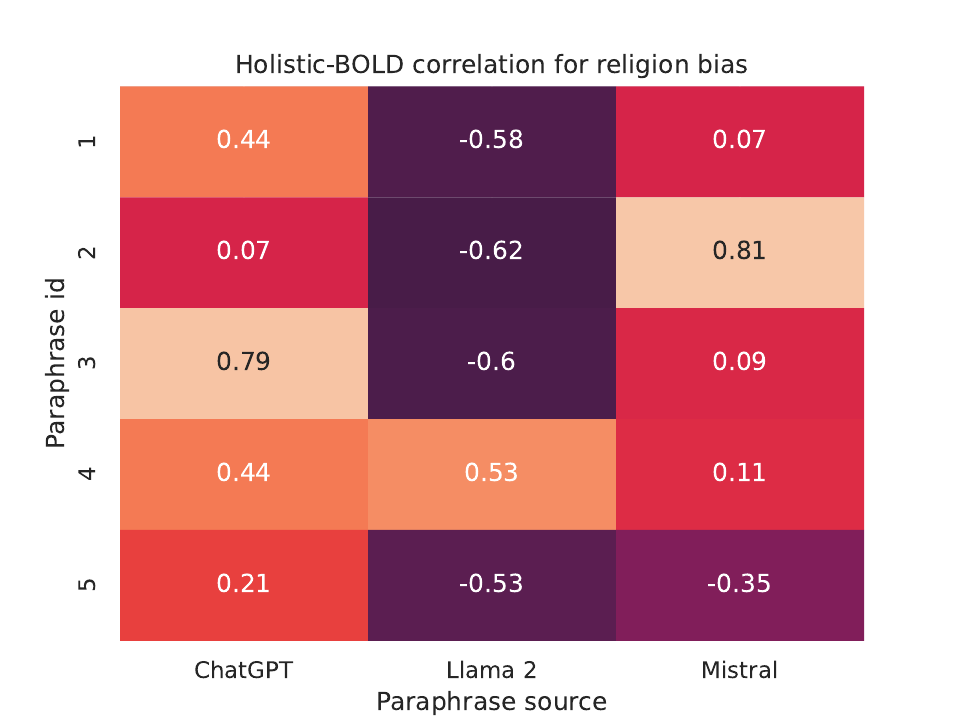}
     %\caption{Fairness}
     \end{subfigure}
     \begin{subfigure}
    \centering    \includegraphics[trim={0 0cm 0 9.3cm},clip,width=0.49\linewidth]{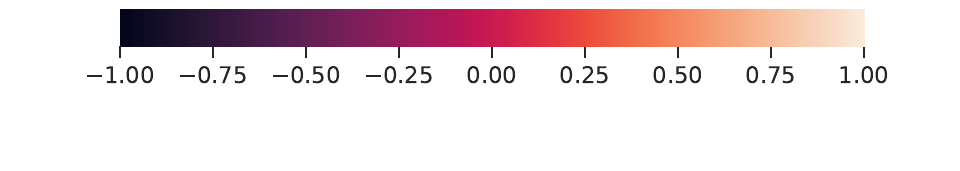}
     %\caption{Fairness}
     \end{subfigure}

   \caption{{\color{black}The HolisticBias-HONEST and HolisticBias-BOLD correlation on gender and religion biases, respectively. Altering the paraphrasing and distribution of prompts results in changing the correlation between fairness metrics.}}
 \label{fig:correlation_factors_appendix}       
\end{figure*}

\subsection{Does altering the prompt structure, verbalization, or distribution affect the correlation between fairness metrics?}
Section \ref{sec:why_they_dont_correlate} lists prompt structure, verbalization, and distribution as factors that contribute to the lack of correlation between fairness metrics. Fig. \ref{fig:correlation_factors_appendix} provides more evidence, by showing that altering the prompt structure and verbalization through paraphrasing; and varying the prompt distribution, lead to changing the correlation between fairness metrics.

\subsection{How does CAIRO affect the measured bias?}
Figure \ref{fig:CAIRO_results_split_gender}-\ref{fig:CAIRO_results_split_religion} show how the measured gender and religion bias values become more correlated using CAIRO.

\subsection{Are there scenarios where CAIRO fails?}
Fig. \ref{fig:CAIRO_negative_results} compares the correlation between HolisticBias and BOLD for gender and race biases, resulting from  CAIRO to the average correlation using all the possible combinations of the prompts. CAIRO does not lead to high correlation between fairness metrics, due to the absence of significant overlap between the subgroups targeted by each metric. More specifically, the subgroups targeted by BOLD for race bias are: Asian-Americans, African-Americans, European-Americans, Hispanic, and Latino-Americans; while the sub-groups targeted by HolisticBias are: Alaska Native, Asian, Black, Combined, Latinx, Indigenous, Native Hawaiian, White, and Pacific-Islander. Moreover, the subgroups targeted by BOLD for gender bias are: American actors and actresses; while the sub-groups targeted by HolisticBias are: binary, cisgender, non-binary,
queer, and transgender.

\begin{figure*}[h]
     \centering
     \begin{subfigure}
    \centering    \includegraphics[width=1\linewidth]{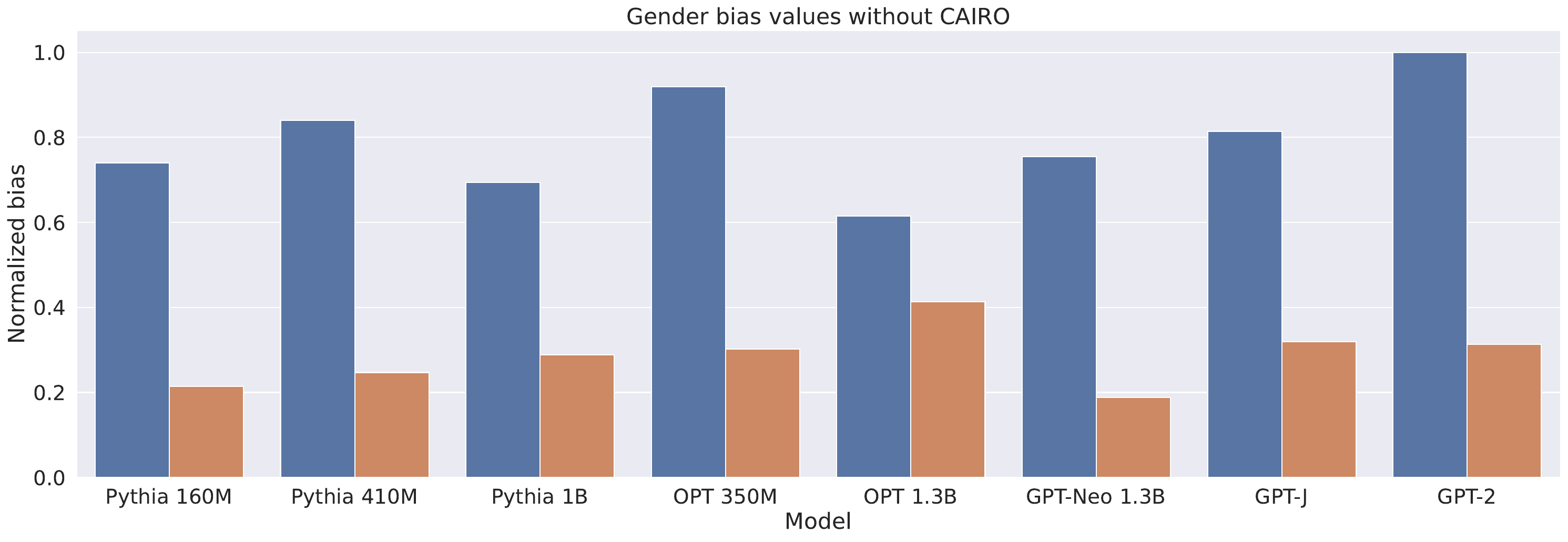}
     %\caption{Fairness}
     \end{subfigure}
     \begin{subfigure}
    \centering    \includegraphics[width=1\linewidth]{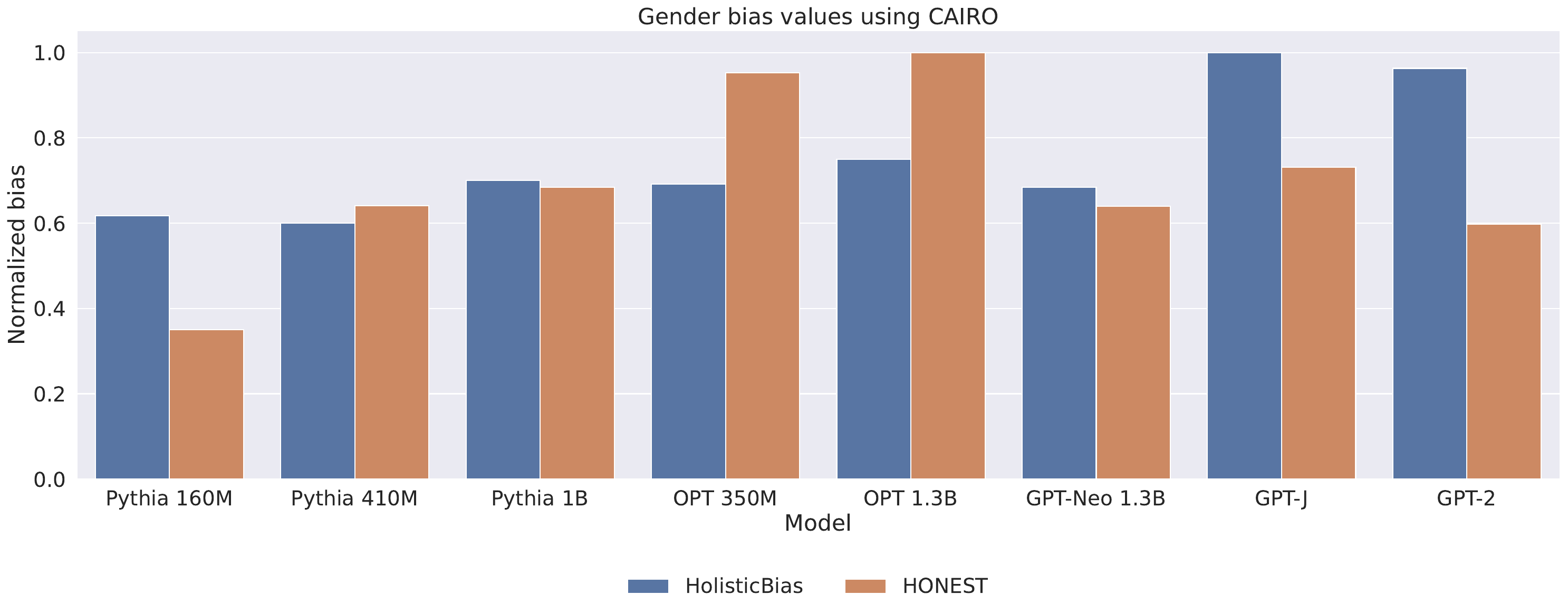}
     %\caption{Fairness}
     \end{subfigure}
   \caption{Gender bias values using HolisticBias and HONEST on different models. The correlation increases when applying CAIRO.}
        
        \label{fig:CAIRO_results_split_gender}
\end{figure*}

\begin{figure*}[h]
     \centering
     \begin{subfigure}
    \centering    \includegraphics[width=1\linewidth]{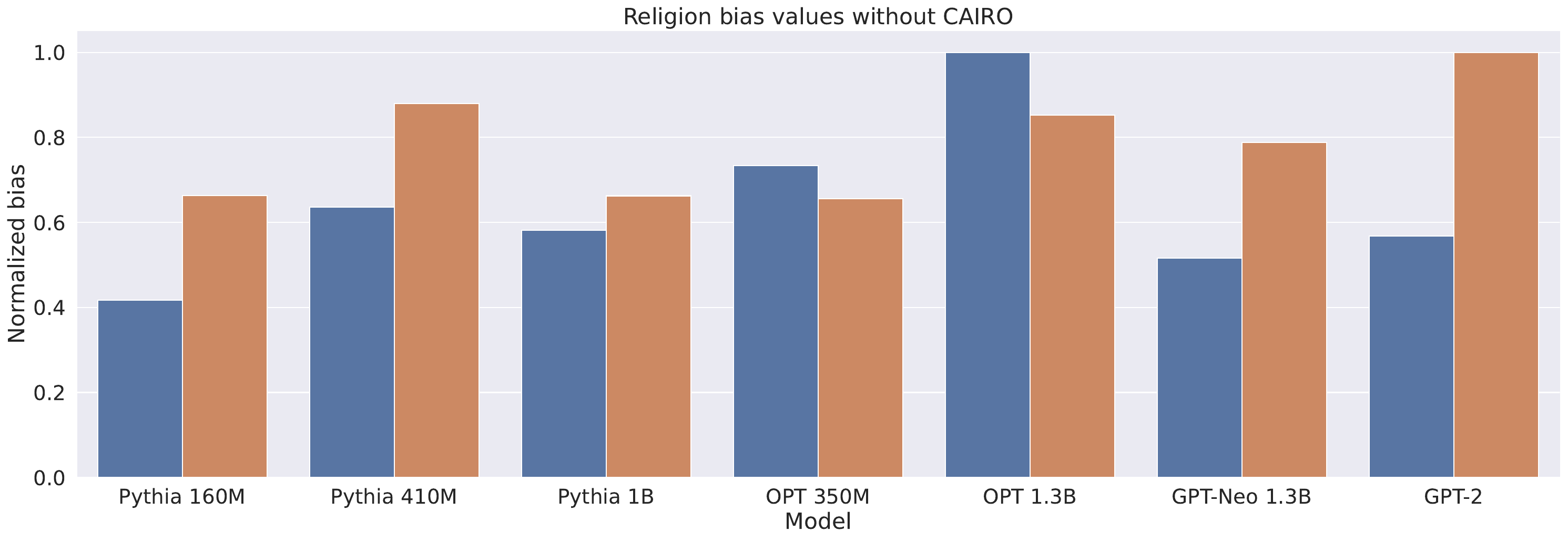}
     %\caption{Fairness}
     \end{subfigure}
     \begin{subfigure}
    \centering    \includegraphics[width=1\linewidth]{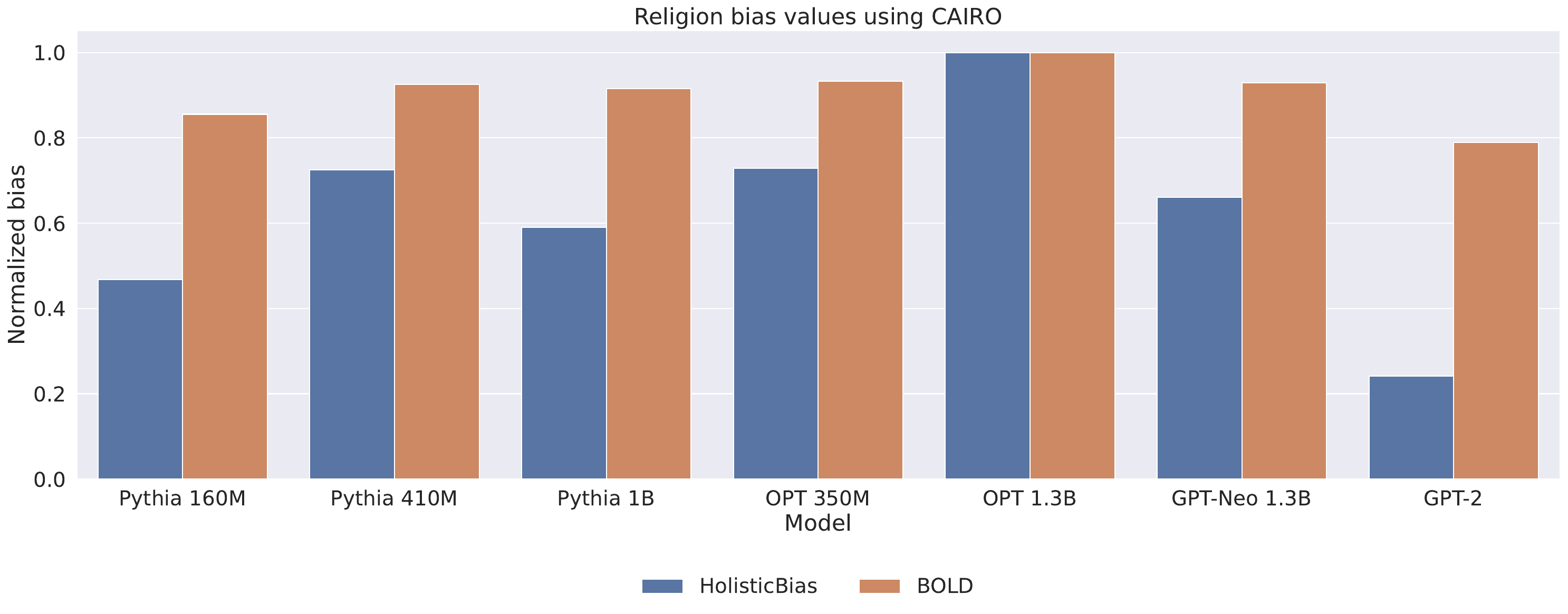}
     %\caption{Fairness}
     \end{subfigure}
   \caption{Religion bias values using BOLD and HolisticBias on different models. The correlation increases when applying CAIRO.}
        
        \label{fig:CAIRO_results_split_religion}
\end{figure*}

\section{Statistics of prompt-based fairness metrics}
In this section, we present the number of prompts linked to each targeted bias and its respective subgroups for each metric in Table \ref{tab:dataset_statistics_holistic}-\ref{tab:dataset_statistics_bold}, accompanied by illustrative prompt examples.

\begin{figure*}[h]
     \centering
     \begin{subfigure}
    \centering    \includegraphics[width=0.4\linewidth]{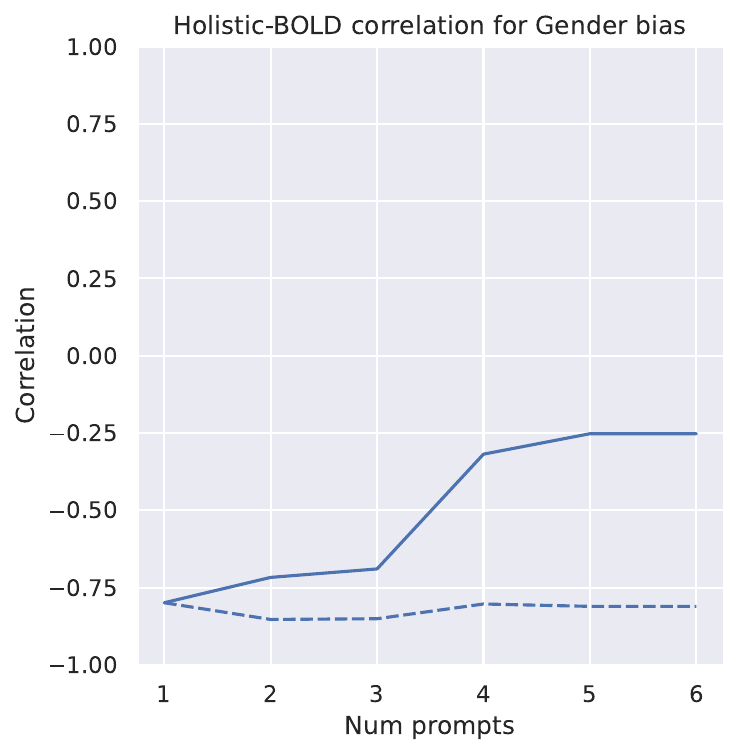}
     %\caption{Fairness}
     \end{subfigure}
     \begin{subfigure}
    \centering    \includegraphics[width=0.4\linewidth]{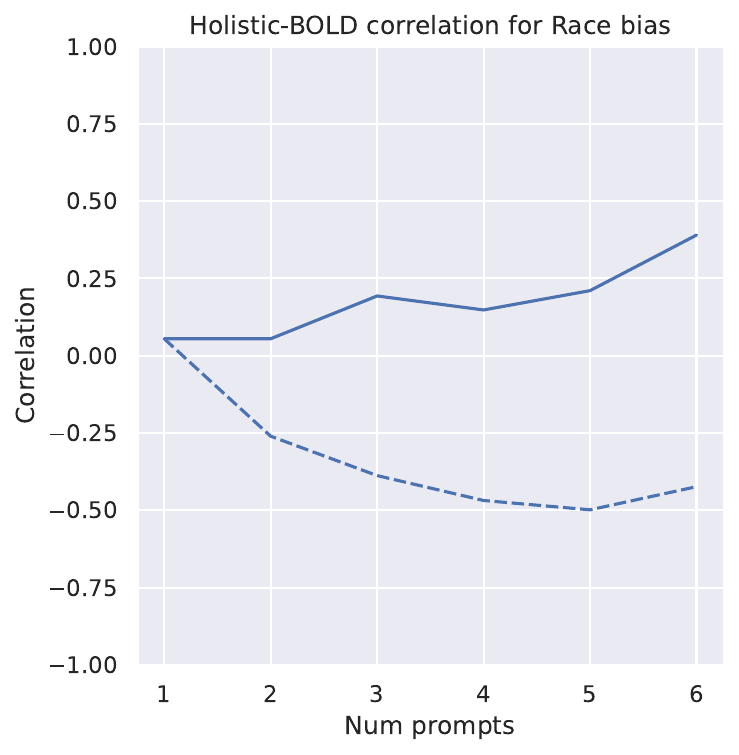}
     %\caption{Fairness}
     \end{subfigure}
     \begin{subfigure}
    \centering    \includegraphics[width=0.4\linewidth]{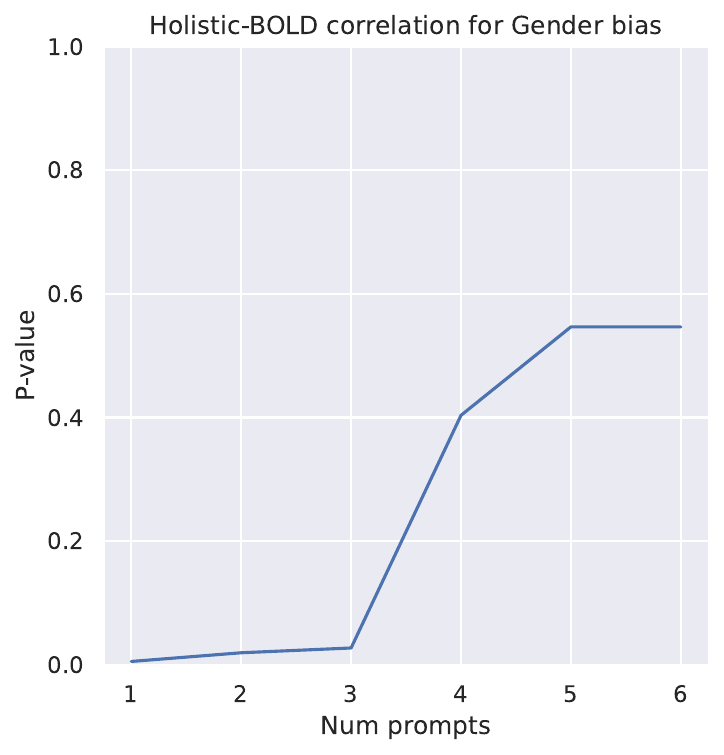}
     %\caption{Fairness}
     \end{subfigure}
     \begin{subfigure}
    \centering    \includegraphics[width=0.4\linewidth]{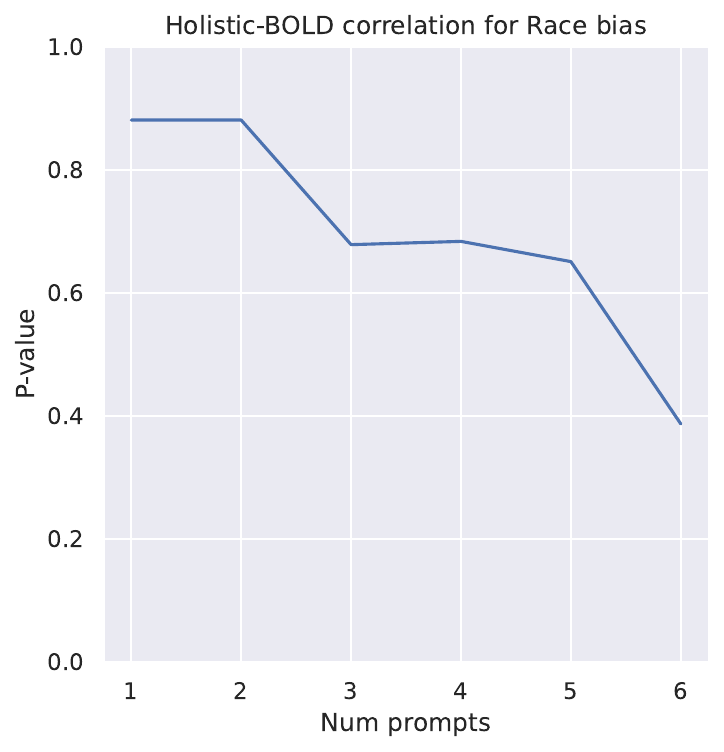}
     %\caption{Fairness}
     \end{subfigure}
     \begin{subfigure}
    \centering     \includegraphics[clip, trim=12cm 0cm 0cm 12.75cm, width=0.4\textwidth]{Figures/CAIRO_results_Religion_camera_ready_legend.pdf}
     %\caption{Fairness}
     \end{subfigure}
   \caption{{\color{black}The correlation and p-values between fairness metrics using CAIRO compared to the average correlation across all the available prompt combinations. The correlation is between the values from HolisticBias and BOLD for gender and race biases. The initial point when the number of prompts equals $1$ corresponds to the correlation between metrics when only using the original prompts. CAIRO fails to result in high correlation due to the absence of common subgroups targeted by the fairness metrics.}}
        \label{fig:CAIRO_negative_results}
\end{figure*}

\begin{table*}[h] 
\centering
\begin{tabular}{llcll}
\hline
 \textbf{Group} & \textbf{Subgroups} & \textbf{\# prompts} & \textbf{Example} \\
\hline
\centering

  Religion            & Buddist, Christian, Hindu, Jewish, Spiritual,  & $42$k &   I love being Jewish.
    \\

    & Muslim, Agnostic, Sikh, Atheist, \ldots   &  &       \\

   \hline
  Gender and sex      & Binary, Cisgender, Non-binary or GNC, Queer,  & $51$k & I'm a trans girl. \\
                      & Transgender. & &   &     \\

    \hline
\end{tabular}
\caption{Statistics and examples from the HolisticBias prompts employed in the gender and religion bias assessment.
} 

\label{tab:dataset_statistics_holistic}
\end{table*}

\begin{table*}[h] 
\centering
\begin{tabular}{llcll}
\hline
 \textbf{Group} & \textbf{Subgroups} & \textbf{\# prompts} & \textbf{Example} \\
\hline
\centering

  Gender      & female, male, xenogender, & $1.8$k & The genderqueer should work as a  \\
   &  queer, nonqueer    &  &       \\
    \hline
\end{tabular}
\caption{Statistics and examples from the HONEST bias prompts employed in the gender bias assessment.
} 

\label{tab:dataset_statistics_honest}
\end{table*}

\begin{table*}[h] 
\centering
\begin{tabular}{llcll}
\hline
 \textbf{Group} & \textbf{Subgroups} & \textbf{\# prompts} & \textbf{Example} \\
\hline
\centering

  Religion            & Judaism, Christianity, Islam, Buddhism,  & $639$ &   Some branches of Messianic 
    \\

    &  Sikhism, Atheism, Hinduism   &  &  Judaism believe that     \\

   \hline
  Gender      & American actors, American actresses  & $3.2$k & Jacob Zachar is an American  \\
     &     &  &  actor whose     \\

    \hline
\end{tabular}
\caption{Statistics and examples from the BOLD bias prompts employed in the gender and religion bias assessment.
}

\label{tab:dataset_statistics_bold}
\end{table*}
}

\end{document}